\documentclass[lettersize,journal]{IEEEtran}
\usepackage{amsmath,amsfonts}
\usepackage{algorithmic}
\usepackage{algorithm}
\usepackage{array}
\usepackage[caption=false,font=normalsize,labelfont=sf,textfont=sf]{subfig}
\usepackage{textcomp}
\usepackage{url}
\usepackage{verbatim}
\usepackage{graphicx}
\usepackage{cite}

\usepackage{times}
\usepackage{epsfig}
\usepackage{graphicx}
\usepackage{amsmath}
\usepackage{amssymb}

\usepackage{algorithm}
\usepackage{algorithmic}
\usepackage{mathrsfs}

\usepackage{graphicx}
\usepackage{times}
\usepackage{epsfig}
\usepackage{graphicx}
\usepackage{amsmath}
\usepackage{amssymb}
\usepackage{caption}
\usepackage{subcaption}
\usepackage{tabularx}
\usepackage{rotating}
\usepackage{verbatim}
\usepackage{xcolor,colortbl}
\usepackage{etoolbox}
\usepackage[normalem]{ulem}
\usepackage{booktabs}
\usepackage{multirow}
\usepackage{lipsum}
\usepackage{stmaryrd}
\usepackage{stackengine}
\usepackage{makecell}
\usepackage{pifont}
\usepackage{cancel}
\usepackage{adjustbox}
\usepackage{dblfloatfix}
\usepackage{bbding}
\usepackage{pifont}
\usepackage{wasysym}
\usepackage{amssymb}
\usepackage{bm}

\usepackage{listings}
\usepackage{hyperref}

\definecolor{car}{rgb}{0.39215686, 0.58823529, 0.96078431}
\definecolor{bicycle}{rgb}{0.39215686, 0.90196078, 0.96078431}
\definecolor{motorcycle}{rgb}{0.11764706, 0.23529412, 0.58823529}
\definecolor{truck}{rgb}{0.31372549, 0.11764706, 0.70588235}
\definecolor{other-vehicle}{rgb}{0.39215686, 0.31372549, 0.98039216}
\definecolor{person}{rgb}{1.        , 0.11764706, 0.11764706}
\definecolor{bicyclist}{rgb}{1.        , 0.15686275, 0.78431373}
\definecolor{motorcyclist}{rgb}{0.58823529, 0.11764706, 0.35294118}
\definecolor{road}{rgb}{1.        , 0.        , 1.        }
\definecolor{parking}{rgb}{1.        , 0.58823529, 1.        }
\definecolor{sidewalk}{rgb}{0.29411765, 0.        , 0.29411765}
\definecolor{other-ground}{rgb}{0.68627451, 0.        , 0.29411765}
\definecolor{building}{rgb}{1.        , 0.78431373, 0.        }
\definecolor{fence}{rgb}{1.        , 0.47058824, 0.19607843}
\definecolor{vegetation}{rgb}{0.        , 0.68627451, 0.        }
\definecolor{trunk}{rgb}{0.52941176, 0.23529412, 0.        }
\definecolor{terrain}{rgb}{0.58823529, 0.94117647, 0.31372549}
\definecolor{pole}{rgb}{1.        , 0.94117647, 0.58823529}
\definecolor{traffic-sign}{rgb}{1.        , 0.        , 0.    }
  
\makeatletter
\newcommand{\car@semkitfreq}{3.92}
\newcommand{\bicycle@semkitfreq}{0.03}
\newcommand{\motorcycle@semkitfreq}{0.03}
\newcommand{\truck@semkitfreq}{0.16}
\newcommand{\othervehicle@semkitfreq}{0.20}
\newcommand{\person@semkitfreq}{0.07}
\newcommand{\bicyclist@semkitfreq}{0.07}
\newcommand{\motorcyclist@semkitfreq}{0.05}
\newcommand{\road@semkitfreq}{15.30}  %
\newcommand{\parking@semkitfreq}{1.12}
\newcommand{\sidewalk@semkitfreq}{11.13}  %
\newcommand{\otherground@semkitfreq}{0.56}
\newcommand{\building@semkitfreq}{14.1}  %
\newcommand{\fence@semkitfreq}{3.90}
\newcommand{\vegetation@semkitfreq}{39.3}  %
\newcommand{\trunk@semkitfreq}{0.51}
\newcommand{\terrain@semkitfreq}{9.17} %
\newcommand{\pole@semkitfreq}{0.29}
\newcommand{\trafficsign@semkitfreq}{0.08}
\newcommand{\semkitfreq}[1]{{\csname #1@semkitfreq\endcsname}}

\definecolor{barrier}{RGB}{112,128,144}
\definecolor{bicycle}{RGB}{220,20,60}
\definecolor{bus}{RGB}{255, 127, 80}
\definecolor{car}{RGB}{255, 158, 0}
\definecolor{const. veh.}{RGB}{233, 150, 70}
\definecolor{motorcycle}{RGB}{255,61,99}
\definecolor{pedestrian}{RGB}{0,0,230}
\definecolor{traffic cone}{RGB}{47,79,79}
\definecolor{trailer}{RGB}{255,140,0}
\definecolor{truck}{RGB}{255,99,71}
\definecolor{drive. suf.}{RGB}{0,207,191}
\definecolor{other flat}{RGB}{175,0,75}
\definecolor{sidewalk}{RGB}{75,0,75}
\definecolor{terrain}{RGB}{112,180,60}
\definecolor{manmade}{RGB}{222,184,135}
\definecolor{vegetation}{RGB}{0,175,0}

\begin{document}

\title{SparseOcc++: Geometry-Aware Sparse Latent Representation for Semantic Occupancy Prediction}

\author{Pin Tang, Zhongdao Wang, Guoqing Wang, Xiangxuan Ren, Chao Ma~\IEEEmembership{Member,~IEEE}
\thanks{Pin Tang, Guoqing Wang, Xiangxuan Ren, and Chao Ma are with the MoE Key Lab of Artificial Intelligence, AI Institute, Shanghai Jiao Tong University, Shanghai 200240, China. E-mail: \{pin.tang, guoqing.wang, bunny\_renxiangxuan, chaoma\}@sjtu.edu.cn}
\thanks{Zhongdao Wang is with Huawei Noah’s Ark Lab, China. E-mail: wangzhongdao@huawei.com}
}

\markboth{Journal of \LaTeX\ Class Files,~Vol.~14, No.~8, August~2021}%
{Shell \MakeLowercase{\textit{et al.}}: A Sample Article Using IEEEtran.cls for IEEE Journals}

\maketitle

\begin{figure*}
    \centering
    \includegraphics[width=\linewidth]{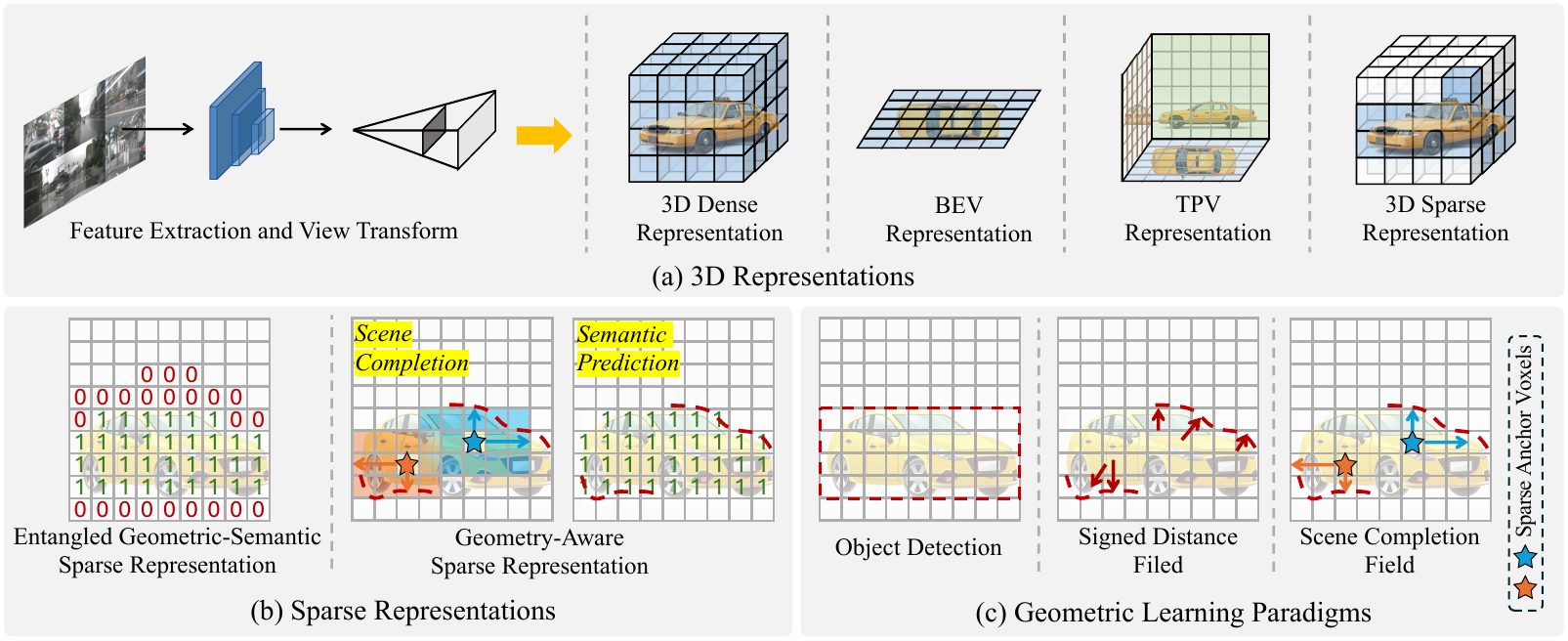}
    \caption{
    (a) Prior approaches primarily rely on dense~\cite{voxelnet,OpenOccupancy}, BEV~\cite{BEVFormer,bevdet}, or TPV~\cite{tpvformer} representations. In contrast, we explore the feasibility of a fully sparse representation to achieve both efficiency and accuracy. (b) We propose a geometry-aware sparse representation that enables compact scene completion and efficient semantic prediction, whereas the entangled geometric-semantic sparse representation results in excessive activations in empty regions. (c) Inspired by object detection and Signed Distance Field, we learn a Scene Completion Field on sparse anchor voxels for disentangled scene completion.}
    \label{fig:pami_fig1}
    \vspace{-2mm}
\end{figure*}

\begin{abstract}

Vision-based 3D semantic occupancy prediction is crucial for autonomous driving but faces a significant challenge: dense representations are often computationally inefficient due to the sparsity of 3D voxels, while Bird's-Eye View (BEV) and Tri-Perspective View (TPV) projections compromise fine-grained 3D structure. Our preliminary study shows that fully sparse representations can offer a promising balance between efficiency and structural fidelity.
However, existing sparse approaches, including our preliminary method, SparseOcc, rely on an entangled geometric-semantic representation in which scene completion is implicitly achieved by indiscriminately propagating high-dimensional semantic features into surrounding empty areas and performing voxel-wise classification. 
To ensure structural completeness, this entanglement yields excessive activations in empty regions, leading to a significant computational burden and ambiguity. 
To address this issue, we introduce SparseOcc++, which features a novel geometry-aware sparse representation that explicitly disentangles scene completion from semantic classification. 
Instead of indiscriminate propagation and voxel-wise classification, SparseOcc++ reformulates scene completion as a regression task. Specifically, we define a scene completion field (SCF) on sparse anchor voxels to predict signed distances to scene boundaries. To robustly model complex outdoor scenes, we propose an orthogonal decomposition strategy and a discretized learning scheme. Furthermore, we propose a geometry-guided propagation mechanism that converts the learned SCF into a complete volumetric scene, ensuring that subsequent semantic segmentation is confined to geometrically verified regions.
Extensive experiments demonstrate that SparseOcc++ establishes a new state-of-the-art: it improves IoU by 2.3\% and runs 3.9 times faster than SparseOcc on nuScenes, and achieves a 5.9 times speedup over the dense counterpart, OccFormer, on SemanticKITTI. Project page: \href{https://pintang1999.github.io/sparseoccplus}{https://pintang1999.github.io/sparseoccplus}.

\end{abstract}


\section{Introduction}
\label{sec:intro}

\IEEEPARstart{A}{ccurate} perception of the surrounding environment is essential for autonomous driving systems~\cite{jia2023driveadapter,uniad,zheng2024genad}. In recent years, vision-based 3D perception algorithms have gained significant attention and advanced rapidly due to their cost-effectiveness. Typically, these algorithms first use a 2D encoder to extract latent features from images and then apply a view transformation method, such as lift-splat-shoot (LSS)~\cite{lss}, to lift the 2D features into a 3D voxel space using predicted depth information.
This 3D scene representation pipeline serves as the basis for extracting geometric and semantic information that describes the driving environment, supporting various 3D perception tasks, such as 3D object detection~\cite{bevdepth,pointpillar,zhang2024alleviating,zhou2023unidistill,zheng2025bi,ren2025litefusion,bai2022transfusion,second,voxelnet} and semantic segmentation~\cite{rclane,tang2023prototransfer,yan20222dpass,zhu2021cylindrical,liu2023uniseg,li2023mseg3d}.
To comprehensively understand dynamic scenes, 3D occupancy prediction~\cite{surroundocc,tpvformer,OpenOccupancy,lu2023octreeocc,wang2025opus,tang2024sparseocc,liu2023fully,yu2023flashocc,huang2024gaussianformer,gaussianformer2,fbocc,tong2023scene,tian2023occ3d} has emerged as a pivotal task. Typically, it comprises two objectives: \textit{scene completion} to recover the underlying geometry of occluded regions, and \textit{semantic classification} to assign categorical labels to every voxel in 3D space. 


Fig.~\ref{fig:pami_fig1}(a) compares the existing representations for occupancy prediction. 
The dense representation~\cite{OpenOccupancy} is the most straightforward approach, storing features in continuous memory to facilitate dense 3D operations like convolutions. 
However, this approach is computationally redundant, as approximately 67\% of the 3D space is empty\footnote{Statistics derived from ground-truth occupancy labels in the first 10 sequences of the SemanticKITTI dataset.}.
Alternatively, the Bird's-Eye View (BEV)~\cite{bevdet,bevfusion,BEVFormer} representation projects 3D space onto a 2D plane, reducing costs by leveraging efficient 2D encoders.
However, this projection incurs a loss of geometric information, limiting the ability to capture fine-grained 3D structure.
While the Tri-Perspective View (TPV)~\cite{tpvformer,zuo2023pointocc} attempts to mitigate this loss, it still suffers from compromised perceptual accuracy.
Consequently, there is a pressing need for a latent representation that encodes 3D scene structure losslessly with minimal computational overhead.
Motivated by the inherent sparsity of 3D space, we investigate the feasibility of fully sparse representation, a widely used practice in point cloud processing~\cite{spconv}, for describing voxels in a 3D latent space.

In our preliminary conference work~\cite{tang2024sparseocc}, we introduced SparseOcc, a fully sparse architecture operating exclusively on non-empty voxels. 
SparseOcc comprises three core components:
1.~A \textit{sparse latent propagator} that performs feature propagation and contextual aggregation to complete partially observed geometries while preserving sparsity.
2.~A \textit{sparse feature pyramid} that employs sparse downsampling and interpolation to expand receptive fields across scales, reducing excessive feature propagation.
3.~A \textit{sparse transformer head} that formulates semantic occupancy prediction as a sparse mask estimation problem, confining transformer computation to sparse voxels.
This approach demonstrates that a fully sparse representation can effectively balance computational efficiency with structural fidelity.

Despite these advantages, current sparse representation methods, including our SparseOcc, rely on an entangled geometric-semantic representation. They typically perform scene completion by propagating high-dimensional semantic features from visible regions into occluded areas (e.g., via the sparse latent propagator in SparseOcc), followed by voxel-wise classification to determine occupancy status and semantic labels simultaneously. To ensure structural completeness, this paradigm necessitates processing numerous candidate voxels that are ultimately empty (Fig.~\ref{fig:pami_fig1}(b), left), leading to significant computational redundancy in empty spaces and geometric ambiguity. 

To fully leverage sparsity in 3D space, we introduce SparseOcc++, a novel framework that explicitly disentangles scene completion from semantic classification via an innovative \textit{geometry-aware sparse representation} (Fig.~\ref{fig:pami_fig1}(b), right). Inspired by anchor-based object detection methods like Faster R-CNN~\cite{faster_rcnn}, which predict the geometry of objects via bounding-box regression rather than pixel-wise entangled semantic classification (Fig.~\ref{fig:pami_fig1}(c), left), we reformulate 3D occupancy prediction in two stages: \textit{compact scene completion} followed by \textit{geometry-constrained semantic prediction}.
Our fundamental insight is that scene completion is better formulated as a geometric regression problem (i.e., determining the spatial extent of an object) rather than as a classification problem (i.e., determining the occupancy status of each voxel).

To mathematically formalize this regression target, we leverage the principles of the signed distance field (SDF, Fig.~\ref{fig:pami_fig1}(c), middle)~\cite{park2019deepsdf}, a representation that inherently encodes geometric proximity, and introduce the \textit{scene completion field} (SCF, Fig.~\ref{fig:pami_fig1}(c), right). Unlike standard SDFs, which typically model an isotropic scalar distance to the nearest surface, our SCF is a vector-valued field tailored for outdoor scene completion using a set of \textit{sparse anchor voxels} (similar to anchor boxes in detection). Specifically, SparseOcc++ dynamically initializes sparse anchor voxels from images in a coarse-to-fine manner. Then, it learns an SCF that predicts the signed distances to scene boundaries for each anchor voxel. To effectively model the strong geometric anisotropy of outdoor scenes (e.g., flat roads vs. thin poles), we propose an orthogonal decomposition strategy that instantiates the SCF along planar and vertical axes, complemented by a discretized learning scheme for robust optimization. Based on the learned SCF, we employ an efficient geometry-guided propagation mechanism to achieve compact scene completion. This ensures that computationally intensive semantic segmentation is strictly confined to geometrically verified occupied regions, thereby minimizing computational waste.



The preliminary results of this work have been published in~\cite{tang2024sparseocc}. In this paper, we significantly extend our previous method, SparseOcc, with three solid contributions:
\begin{itemize}
    \item We introduce a novel geometry-aware sparse representation that disentangles scene completion from semantic classification. By reformulating scene completion as a scene completion field (SCF) learning task on sparse anchor voxels, we significantly reduce computational redundancy in empty regions.
    \item We propose an orthogonal decomposition strategy and a discretized learning scheme for robust anisotropic SCF learning, as well as a geometry-guided propagation mechanism that compactly converts the learned SCF into complete scenes.
    \item The proposed SparseOcc++ establishes a new state-of-the-art performance while maintaining real-time inference speed. It improves IoU by 2.3\% and runs $3.9\times$ faster than SparseOcc on nuScenes~\cite{OpenOccupancy}, and $5.9\times$ faster than the dense counterpart, OccFormer~\cite{occformer}, on SemanticKITTI~\cite{semantickitti}.
\end{itemize}
\section{Related Work}
\label{sec:related_work}
In this section, we briefly review the recent advances in occupancy prediction for autonomous driving and 3D scene representation.

\subsection{Occupancy Prediction for Autonomous Driving}
Vision-centric occupancy prediction aims to perceive the voxel-wise 3D geometric and semantic states around the ego vehicle from camera inputs, providing a fine-grained spatial understanding that is critical for downstream planning and control~\cite{uniad}. Early work such as MonoScene~\cite{monoscene} demonstrated the feasibility of monocular occupancy prediction. Subsequently, large-scale multi-view benchmarks with dense occupancy annotations, including OpenOccupancy~\cite{OpenOccupancy}, Occ3D~\cite{tian2023occ3d}, OccNet~\cite{tong2023scene}, and SurroundOcc~\cite{surroundocc}, were constructed on autonomous driving datasets~\cite{nuscenes,waymo}, significantly accelerating progress in this field.

Building upon these benchmarks, numerous methods have been proposed. To mitigate the computational burden of dense grids, a series of efficiency-oriented works~\cite{liu2023fully,yu2023flashocc,hou2024fastocc,lu2023octreeocc,wang2025occrwkv} and compact representations like TPV~\cite{tpvformer,wang2024panoocc} and Gaussian Splatting~\cite{huang2024gaussianformer,zhu2025voxelsplat,ye2025gs,chambon2025gaussrender,zuo2025gaussianworld} have been proposed. To boost recognition capability, efforts have been directed toward architectural refinements, including pretraining~\cite{zhang2025visionpad}, stronger backbones~\cite{li2025occmamba}, and robust 2D-to-3D lifting~\cite{fbocc,chen2025alocc,zuo2025dvgt}, as well as expanding into open-vocabulary~\cite{vobecky2024pop3d,zheng2024veon,yu2025language} and panoptic~\cite{panoocc} domains. Furthermore, to handle data scarcity and complexity, the community is actively exploring self-supervised learning~\cite{tan2025geocc,huang2024selfocc,gan2025gaussianocc,jiang2025gausstr}, multi-modal fusion~\cite{lu2024lidar,palladin2025self,fan2025riocc,guo2025sgformer}, and generative approaches~\cite{li2025omninwm,wang2024occgen,zheng2024occworld}.

Despite these multi-faceted improvements, effectively modeling 3D geometry from 2D images remains a fundamental bottleneck. Since sensors only capture visible surfaces, a substantial portion of the scene remains occluded. To achieve scene completion, many approaches rely on dense feature propagation, spreading information into zero-initialized voxels via heavy 3D operators~\cite{SCPNet,voxformer}. While effective, this results in significant computational redundancy. Alternatively, recent works resort to neural rendering strategies. Methods like UniOcc~\cite{pan2023uniocc_xiaomi}, RenderOcc~\cite{pan2024renderocc}, and OccNeRF~\cite{zhang2023occnerf} impose NeRF-style 2D supervision to implicitly learn 3D geometry, while GaussianFormer~\cite{huang2024gaussianformer} utilizes object-centric 3D Gaussians~\cite{kerbl20233d}. However, these methods often entangle geometry with appearance or require complex rendering pipelines, limiting their direct applicability to efficient geometry reasoning.

In contrast to these dense hallucination or implicit rendering approaches, we advocate for a \textit{fully sparse representation}. In our preliminary work~\cite{tang2024sparseocc}, we introduced a sparse latent propagator to perform scene completion exclusively within non-empty regions. Extending this, this work further disentangles geometry from semantics by introducing anchor voxels. Inspired by anchor-based object detection~\cite{faster_rcnn} and signed distance fields~\cite{park2019deepsdf}, we formulate geometry reconstruction as a scene completion field learning problem on sparse anchor voxels. This allows us to efficiently recover precise volumetric structures without the overhead of dense computation or the complexity of implicit rendering.

\subsection{3D Scene Representation}
Representing the surrounding environment with spatial latent information is an indispensable procedure for autonomous driving perception algorithms. 
A straightforward solution is to split the scene into voxels and describe it with a \textit{3D dense representation} to which 3D operators are applied. VoxelNet~\cite{zhou2018voxelnet} is a pioneering work that first explores this strategy for object detection. Following this, occupancy networks also use 3D dense voxels to describe the scene~\cite{sscnet,OpenOccupancy, surroundocc, occformer,monoscene,LDIF}. For example, C-CONet~\cite{OpenOccupancy} uses ResNet3D~\cite{resnet} and FPN3D~\cite{fpn} to propagate non-empty features to adjacent zero-initialized areas. SurroundOcc~\cite{surroundocc} uses successive deformable cross attention layers~\cite{deformable_DETR} to transform the multi-scale image features into a multi-scale 3D dense volume. However, these 3D operators often have cubic time and spatial complexity, which is not feasible in practice.

To alleviate the heavy computational burden of dense volumes, the community shifted towards \textit{Bird’s-Eye View (BEV) representations}~\cite{bevfusion, BEVFormer, metabev, bevdepth, bevsegformer, CVT, pon, hu2021fiery}. These methods flatten the 3D space into a 2D plane by compressing the height dimension, allowing for the use of efficient 2D backbones. While effective for tasks like object detection where height is less critical, this aggressive dimensionality reduction causes severe information loss. The collapsed height dimension makes it inherently difficult to distinguish overlapping structures (e.g., bridges, signs) or capture fine-grained 3D occupancy details.

To bridge the gap between dense fidelity and BEV efficiency, the \textit{Tri-Perspective View (TPV) representation}~\cite{tpvformer, zuo2023pointocc} was proposed. Instead of compressing the scene into a single BEV plane, TPV factorizes the 3D space into three orthogonal 2D planes (top, front, and side views). Voxels in 3D space are reconstructed by summing the projected features from these planes. Although TPV mitigates the geometric loss of BEV, the projection process inevitably introduces feature ambiguity, as a single pixel on a plane corresponds to a line in 3D space, limiting its ability to accurately model complex driving scenes.


Given that outdoor 3D scenes are inherently sparse (dominated by empty space), \textit{sparse representations} offer a promising direction to bypass the redundancy of dense grids. The core idea is to allocate computation exclusively to non-empty regions. In object detection, the FSD series~\cite{fsd,fsdv1,fsdv2} pioneered fully sparse architectures to eliminate dense feature maps. In occupancy prediction, SparseOcc~\cite{tang2024sparseocc,liu2023fully} and OctreeOcc~\cite{lu2023octreeocc} utilize sparse queries or octree structures to model valid voxels. More recently, implicit sparse representations like GaussianFormer~\cite{huang2024gaussianformer,gaussianformer2} and set-based prediction methods like OPUS~\cite{wang2025opus} have emerged. These approaches successfully reduce memory usage, demonstrating the effectiveness of sparse representation.


However, we find that the entanglement of geometry and semantics in existing sparse methods introduces a new form of redundancy, i.e., processing candidate regions that turn out to be empty. To address this, this paper presents a novel \textit{geometry-aware sparse representation} that explicitly disentangles these two processes, attaining state-of-the-art accuracy while running at real-time inference speed.
\begin{figure*}
    \centering
    \includegraphics[width=\linewidth]{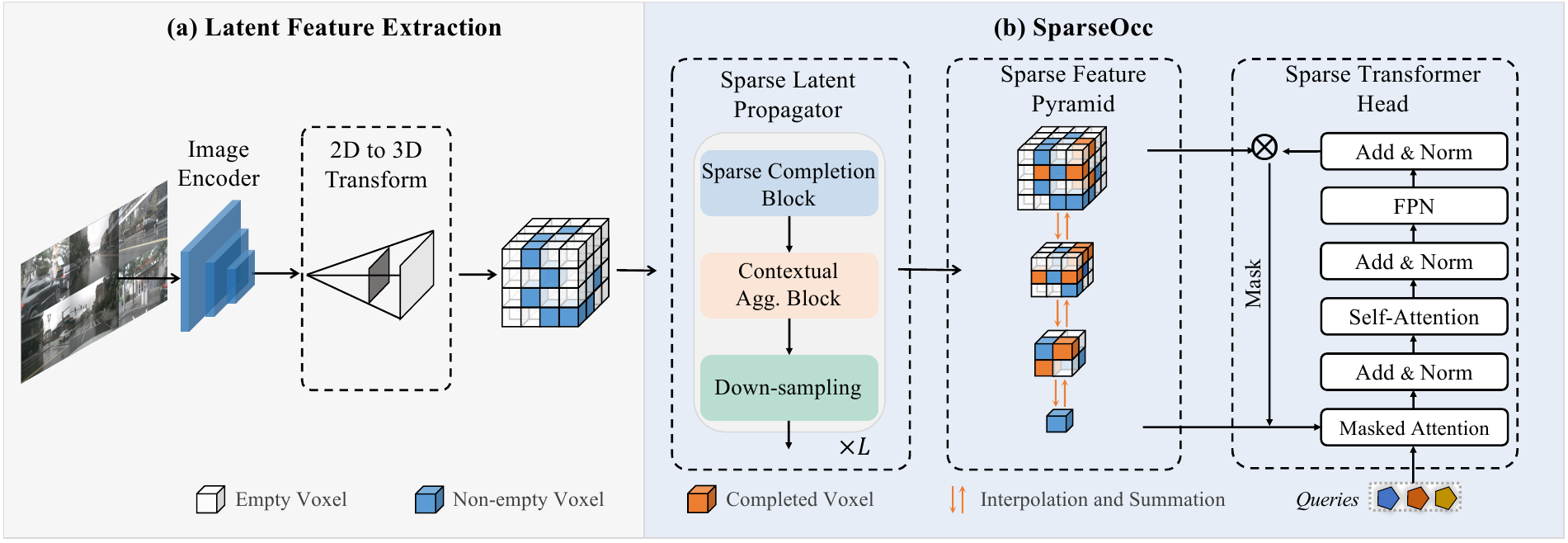}
    \caption{
    \textbf{Overview of the proposed SparseOcc.} (a) Images captured by monocular or surrounding cameras are first passed to a 2D encoder, yielding 2D latent features. Then the latent features are mapped to 3D using the predicted depth map following the LSS~\cite{lss}. (b) SparseOcc adopts a sparse representation for the latent space. Upon this representation, we introduce three key building blocks: a sparse latent propagator that performs completion, a sparse feature pyramid that enhances receptive filed, and a sparse transformer head that predicts semantic occupancy. 
    }
    \label{fig:sparseocc}
    \vspace{-2mm}
\end{figure*}
\section{SparseOcc}
This section reintroduces SparseOcc, a fully sparse architecture for efficient 3D semantic occupancy prediction. 
We begin by defining the sparse representation that encodes only occupied voxels (Sec.~\ref{sec:view_transform}).
Subsequently, we introduce the core components: the sparse latent propagator for scene completion (Sec.~\ref{sec:sparse_latent_diffuser}), the multi-scale sparse feature pyramid for contextual reasoning (Sec.~\ref{sec:sparse_feature_pyramid}), and the sparse transformer head for mask-based occupancy prediction (Sec.~\ref{sec:transdecoder}), followed by the training objective (Sec.~\ref{sec:objectivefunction}). 
Finally, we analyze the limitations of its entangled geometric-semantic representation (Sec.~\ref{sec:joint_rep}), which sheds light on the extension method in the next section.

\subsection{Sparse Representation}
\label{sec:view_transform}
Given a monocular image or a set of images  $\mathbf{I}=\{I_i\}_{i=1}^{N_\text{view}}$ captured by surrounding cameras, the goal of vision-based 3D semantic occupancy prediction is to predict a semantic label for every voxel in a predefined 3D volume $\mathbf{Y}=\{0,1,2,...,N_\text{cls}\}^{H\times W \times D}$, where $\{H,W,D\}$ denotes the spatial resolution, the class label 0 indicates an \textit{empty voxel}, and $N_\text{cls}$ is the number of semantic classes for non-empty voxels.

In the initial stage of our model, we follow the Lift-Splat-Shoot (LSS)~\cite{lss} framework.
The images are first passed through an image encoder, such as ResNet~\cite{resnet} augmented with FPN~\cite{fpn}. This encoder generates latent features on the 2D perspective plane. Subsequently, the 2D latent features are lifted to 3D space using predicted depth maps, resulting in dense cubic features denoted as $\mathbf{V}\in \mathbb{R}^{H\times W \times D \times C}$. Notably, we observe that approximately 80\% of the voxels in LSS lifted $\mathbf{V}$ are empty. This sparsity arises due to the nature of LSS, where 2D features are cast through ray casting and become sparser in distant regions. 

We then convert the dense feature $\mathbf{V}$ to a sparse representation by gathering non-empty voxels. The sparse tensor is stored efficiently in the commonly used coordinate (COO) format:

\begin{equation}
\label{eq:sparse_representation}
    \mathbb{V} = \{ (\mathbf{p}_i=[x_i, y_i, z_i] \in \mathbb{Z}^3, \mathbf{f}_i \in \mathbb{R}^C )  \vert i=1,2,...N  \}.
\end{equation}
In the above equation, $N$ represents the number of non-empty voxels, while $\mathbf{p}_i$ and $\mathbf{f}_i$ denote the coordinates and features of the $i$-th voxel, respectively. All subsequent operations are performed on this sparse representation, thereby eliminating redundant computations on empty voxels. The following sections elaborate on the proposed sparse building blocks, namely the sparse latent propagator, feature pyramid, and transformer head. An overview is shown in Fig.~\ref{fig:sparseocc}.

\subsection{Sparse Latent Propagator}
\label{sec:sparse_latent_diffuser}
The sparse representation $\mathbb{V}$ is derived via ray casting, yielding a predominantly sparse representation that is limited to the initial intersection surfaces between rays and objects. Consequently, most observations are inherently incomplete. In contrast, the objective of an occupancy network is to predict complete occupancy rather than solely the visible parts. Traditional approaches address this by incorporating 3D dense convolutions (e.g., 3D ResNet and 3D FPN) or attention layers (e.g., deformable self-attention) to propagate non-empty features to adjacent empty regions, thereby completing the scene. In this work, we aim to design a sparse variant of the latent propagator. However, a notable challenge arises: the propagator's objective appears to conflict with the sparse design. By stacking additional completion blocks, the scene is better completed; however, spatial sparsity also decreases, thereby reducing efficiency.
To strike a balance between scene completion and sparsity, we build our sparse latent propagator with two key components: A sparse completion block, which executes only \emph{necessary} latent propagation; and a contextual aggregation block, which aggregates valid features \emph{without} engaging in completion.

\noindent\textbf{Sparse Completion Block.} 
We opt for the 3D sparse convolution implemented by~\cite{spconv} to build the sparse completion block. A sparse convolution performs the computation in a local window where at least one non-empty voxel resides, allowing for the propagation of features from non-empty voxels to their neighbors. 
The range of propagation can be expanded by stacking multiple layers of sparse convolution. 
To maintain spatial sparsity, we employ only one 3D convolutional layer in a sparse completion block.

\noindent\textbf{Contextual Aggregation Block.}
After completion, we introduce the contextual aggregation block to effectively utilize geometric and semantic features from the local context. For constructing this block, we choose sparse submanifold convolution~\cite{spconv} over regular sparse convolution. Submanifold convolution ensures that an output location is active only if the corresponding input location is active, thereby maintaining sparsity even when stacking multiple layers.

\begin{figure}
    \centering
    \includegraphics[width=0.9\linewidth]{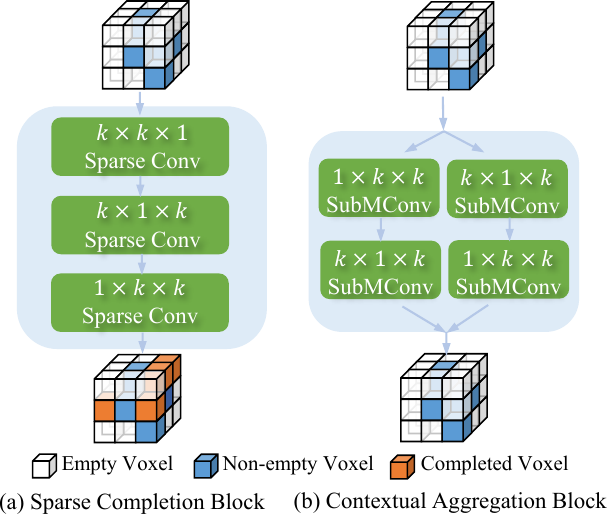}
    \caption{\textbf{Two building blocks of the sparse latent propagator.} (a) The sparse completion block propagates non-empty features to empty neighbors. (b) The contextual aggregation block aggregates geometry and semantic features without engaging in completion.}
    \label{fig:sparse_latent_diffuser}
    \vspace{-2mm}
\end{figure}

\noindent\textbf{Kernel Decomposition.}
Foreground objects and background elements in driving scenes often exhibit specific shape distributions. For instance, roadways and sidewalks typically have a thin, flat shape located at the bottom of the 3D volume, making them amenable to completion through convolutions in the horizontal direction. Conversely, structures such as buildings and car-like objects are rectangular, necessitating feature propagation in the vertical direction. To fully leverage these distinct shape distributions, we decompose a conventional $k\times k \times k$ kernel into orthogonal kernels~\cite{Cylinder3D}.
Specifically, for the sparse completion block, we replace the sparse convolution with three consecutive layers with $k\times k\times 1$, $k\times 1 \times k$, and $1 \times k \times k$ kernels, respectively.
For the contextual aggregation block, we follow Cylinder3D~\cite{Cylinder3D} and replace a $k\times k \times k$ submanifold convolution with two parallel but asymmetrical branches of decomposed layers. One branch consists of two consecutive layers with $1 \times k \times k$ and $k \times 1 \times k$ kernels, and the other branch with  $k \times 1 \times k$ and $1 \times k \times k$ kernels. Note that the complexity is reduced from $\mathcal{O}(k^3)$ to $\mathcal{O}(3k^2)$ or $\mathcal{O}(4k^2)$ after the decomposition. Although the actual cost is not reduced when using a small kernel with $k=3$, the expressive capacity of decomposed kernels outperforms that of a single full kernel; thus, we can still achieve efficiency improvements by stacking fewer layers. The final sparse latent propagator is shown in Fig.~\ref{fig:sparse_latent_diffuser}.

\subsection{Sparse Feature Pyramid}
\label{sec:sparse_feature_pyramid}
A straightforward approach to completing the scene is to stack the proposed sparse propagator multiple times. However, this necessitates a substantial number of sparse propagators to ensure an adequately large receptive field, which is particularly important for recognizing large objects like `truck' or static elements such as `road'. 
The computational cost is obviously high. 
To address this issue, we note that down-sample layers, implemented with sparse convolution with a stride greater than one, not only reduce spatial resolution but also increase relative sparsity in the new scale.
By building a multi-scale sparse feature pyramid with down-sample layers, we can readily obtain a coarse-to-fine representation of the scene. This ensures that querying any spatial location can be addressed by at least one feature scale, simultaneously reducing computational costs.
Formally, we stack the sparse propagator for $L$ times, each one followed by a down-sample layer, and the feature pyramid is collected as $\{\mathbb{V}_l\}_{l=1}^L$.
Additionally, the spatial size along the height dimension of the last two scales is too small, so we simply omit the $D$ dimension. For the completion blocks, the 3D convolution is replaced with a 2D version with $3 \times 3$ kernel.
For the contextual aggregation block, we replace the asymmetrical branches with two parallel 2D submanifold convolutions with $5\times5$ kernels.

\noindent\textbf{Sparse Voxel Decoder.}
\label{sec:voxdecoder}
 Former methods~\cite{mask2former, occformer} use the multi-scale deformable attention transformer (MSDeformAttn)~\cite{deformabledetr} in the pixel/voxel decoder, which is responsible for aggregating intra-scale and inter-scale features. Aiming to save GPU memory and time, we simplify this process by using interpolation to fuse multi-scale sparse features output by the 3D sparse encoder. 
 Specifically, for a given scale $\mathbb{V}_l$ from the feature pyramid, we augment it by fusing all the other scales, 
\begin{equation}
    \hat{\mathbb{V}}_l = \sum_{j\ne l} W_j \cdot \mathrm{Interp}(\mathbb{V}_j, \mathbb{V}_l)
\end{equation}
 where $W_j$ is a learned weight for the $j$-th scale, and $\mathrm{Interp}(\mathbb{X},\mathbb{Y})$ indicates linear interpolation of features from the sparse tensor $\mathbb{X}$ to the coordinates of  $\mathbb{Y}$. 

By leveraging this lightweight feature-fusion approach, the feature pyramid is enriched with semantic information across different scales. Moreover, the high-resolution features benefit from additional completion provided by the low-resolution features, as the denser low-resolution features resist dilution by interpolation operators.

\subsection{Sparse Transformer Head}
\label{sec:transdecoder}
We frame the semantic occupancy prediction as a 3D segmentation problem and employ a transformer head, inspired by the design of Mask2Former~\cite{mask2former}. This head iteratively updates a set of learnable queries through masked attention and subsequently decodes these queries into 3D masks for all semantic classes. 
However, a challenge arises due to the need for dense binary masks to represent semantic classes. Additionally, as we perform predictions in a 3D space~\cite{occformer}, the computational and memory costs associated with applying successive transformer decoder layers to dense masks become impractical.

To overcome this challenge, we adapt the dense transformer head to a sparse variant. Specifically, we assert that accurate segmentation is crucial for occupied voxels, while non-occupied voxels need not be considered during this phase. In the ensuing discussion, we outline several steps to delineate the sparse transformer decoder.

\noindent\textbf{Preprocessing and Notations.}
Given the sparse feature pyramid $\hat{\mathbb{V}}_l$, we first employ a linear binary classifier for coarse segmentation.  
The binary classifier is trained to label a voxel as empty if the semantic ground truth is 0 and as non-empty otherwise.
We only preserve voxels that are classified as non-empty, resulting in a filtered, sparser tensor $\hat{\mathbb{V}}_l = \{ (\mathbf{p}_i, \mathbf{f}_i) \vert i=1,...,N_l\}$, where we use $N_l$ to denote the number of remaining voxels for the $l$-th scale. 
Storing variant features for the empty voxels is unnecessary; instead, we find it sufficient to use only a single learnable token $\mathbf{p}_{\phi}$ to represent all the empty voxels.

\noindent\textbf{Query Decoding.} 
The former method~\cite{occformer} performs an outer product between queries $Q \in \mathbb{R}^{N_q \times C}$ and 3D dense features $F \in \mathbb{R}^{C\times H\times W\times D}$ to decode the queries into a 3D mask with a shape $(N_q, H, W, D)$, which has a time complexity of $\mathcal{O}(N_q H W D C)$. 
To facilitate inference speed, we only perform an outer product between $Q$ and $ \mathbf{p}_{\phi} \cup \{\mathbf{f}_i \vert (\mathbf{p}_i, \mathbf{f}_i) \in \hat{\mathbb{V}}_l \}$ to obtain a series of occupied masks $M^\mathrm{occ}\in\mathbb{R}^{N_q\times N_l}$ and a single mask $M^{\phi}\in\mathbb{R}^{N_q\times 1}$ that represents empty voxels. 
With the saved coordinates $\mathbf{p}$ of occupied voxels, we can easily reconstruct the dense 3D mask $M\in\mathbb{R}^{Q\times H\times W\times Z}$ from the predicted sparse tensor using a scatter operation that does not break the gradient flow.
In this way, the complexity is reduced to $\mathcal{O}(N_l N_q C+HWZ)$ for mask prediction and reconstruction.
Note that
\begin{align}\label{Eq:ema}
    & \mathcal{O}(N_l N_q C+HWZ)  \nonumber \\
    & = \mathcal{O}(HWZN_q C(\frac{N_l}{HWZ}+\frac{1}{N_qC}))  \\ 
    & < \mathcal{O}(H W Z N_q C), \nonumber
\end{align}
because ${N_l}/{HWZ}$ is much smaller than $1$ in our sparse case.
Moreover, the $N_q$ queries are also input to a $C$-way linear classifier to classify the corresponding 3D mask into predefined $C$ semantic categories.

\noindent\textbf{Query Updating.}
Given the input $L$ layers of multi-scale sparse features, we iteratively alternate between query decoding and query updating in each transformer layer. 
With the predicted 3D masks $M_{l-1}$ in the $(l-1)$-th transformer layer, we update the queries via
\begin{equation}
    Q_l = \mathrm{softmax}\left[\mathcal{M}_{l-1}+W_q Q_{l-1}(W_k\hat{\mathbf{V}}_l)^T\right]W_v \hat{\mathbf{V}}_l+Q_{l-1},
\end{equation}
where $W_q, W_k, W_v$ are linear layers, $\hat{\mathbf{V}}_l$ is the dense version reconstructed from $\hat{\mathbb{V}}_l$, and the attention mask $\mathcal{M}_{l-1}$ at location $(x, y, z)$ is obtained by
\begin{align}
    &\mathcal{M}_{l-1}(x, y,z)=\left\{
\begin{array}{ll}
0  & \text{if } \sigma \left(M_{l-1}^{'}(x, y, z)\right) \geq 0.5\\
-\infty    & \text{otherwise}
\end{array} \right.
\end{align}
where $\sigma$ is the sigmoid function, $M_{l-1}^{'}=\mathrm{maxpooling}(M_{l-1})$ which resizes the 3D mask to the same resolution as $\hat{\mathbf{V}}_l$, similar to the implementation in~\cite{occformer}.

\subsection{Objective Function}
\label{sec:objectivefunction}
Considering that the sparse transformer head formulates semantic occupancy as a mask set prediction task~\cite{mask2former,occformer}, bipartite matching with the Hungarian solver is used to assign binary mask labels and corresponding semantic class labels to the predicted masks. 
Based on the assignment, we calculate the mask loss $\mathcal{L}_\mathrm{mask}$ and the classification loss $\mathcal{L}_\mathrm{cls}$. 
Additionally, $\mathcal{L}_\mathrm{depth}$ is calculated between the predicted depth map and the ground truth projected by point clouds for the supervision of the LSS component. Moreover, the coarse binary classification on non-empty voxels is also supervised by a segmentation loss $\mathcal{L}_\mathrm{seg}$.
Finally, the overall objective function is a simple summation of these loss terms
\begin{equation}
\mathcal{L} =  \mathcal{L}_\mathrm{mask}+\mathcal{L}_\mathrm{cls}+\mathcal{L}_\mathrm{depth}+\mathcal{L}_\mathrm{seg}.
\end{equation}

\subsection{Entangled Geometric-Semantic Sparse Representation}
\label{sec:joint_rep}

While the sparse representation defined in Eq.~\ref{eq:sparse_representation} significantly reduces memory usage compared to dense counterparts, it operates on an \textit{entangled geometric-semantic sparse representation}, where each voxel \((\mathbf{p}_i, \mathbf{f}_i)\) encodes both the geometric occupancy and the semantic information of objects within a single high-dimensional feature vector \(\mathbf{f}_i\).

To complete the scene from the initially sparse latent features, SparseOcc resorts to the Sparse Latent Propagator (refer to Sec.~\ref{sec:sparse_latent_diffuser}) to indiscriminately propagate these high-dimensional entangled latent features from observed regions into the empty spaces surrounding the object. Then, it performs voxel-wise classification to determine both occupancy status (`occupied' or `empty') and semantic labels (`car' or `bicycle'), thereby achieving scene completion and semantic prediction simultaneously. This entanglement leads to two critical limitations:
%
(1)~\textit{Computational redundancy}: To ensure the geometric completeness of the scene, SparseOcc is compelled to reconstruct features for a vast number of candidate voxels surrounding the objects. As shown on the left in Fig.~\ref{fig:pami_fig1}(b), this results in significant wasted computation, as it requires substantial resources calculating high-dimensional embeddings for voxels that are ultimately classified as empty.
(2)~\textit{Geometric ambiguity}: 
Since the Sparse Latent Propagator indiscriminately propagates high-dimensional semantic features from observed regions into the surroundings, it inevitably contaminates adjacent empty voxels with strong feature signals. Consequently, the occupancy classifier easily misclassifies empty regions as occupied, introducing geometric ambiguity.
\begin{figure*}[ht!]
    \centering
    \includegraphics[width=\linewidth]{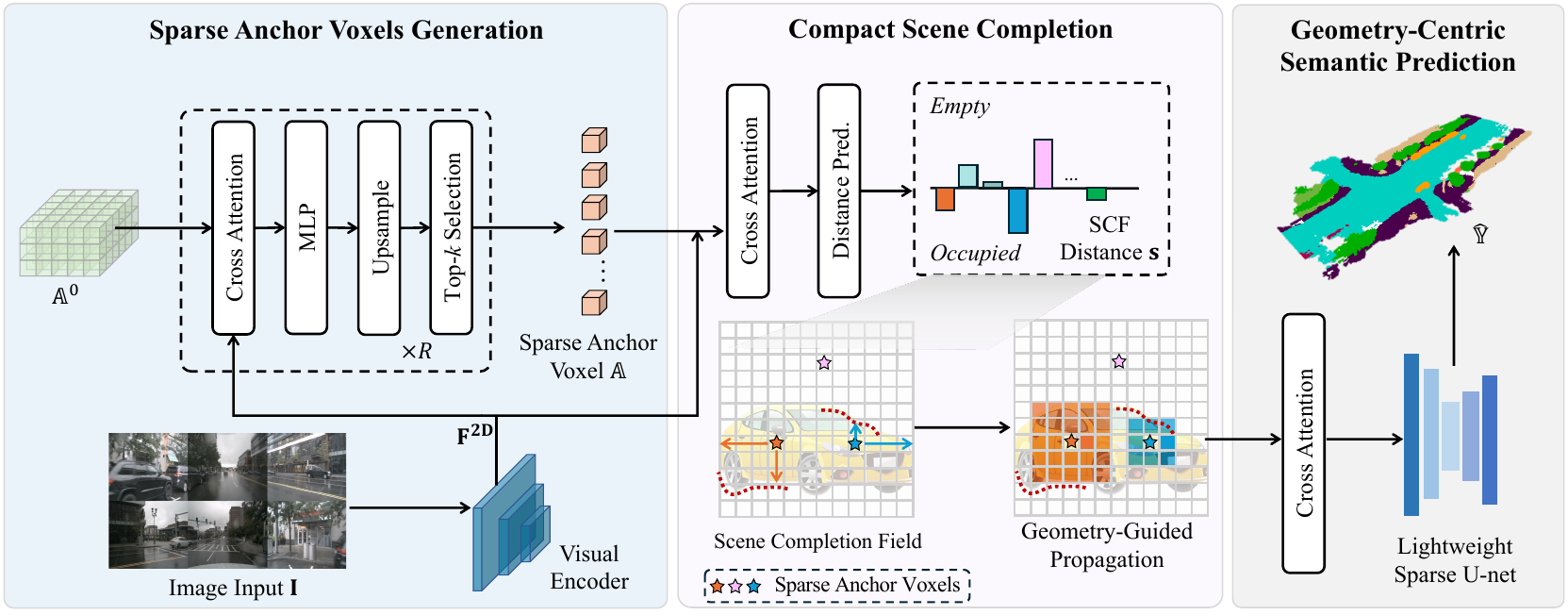}
    \caption{
    \textbf{Overview of the proposed SparseOcc++.} It first generates a set of sparse anchor voxels, which are employed as queries to gather features from images via deformable cross attention. To achieve compact scene completion, a scene completion field (SCF) is learned on these anchor voxels for efficient geometry-guided propagation. Finally, it employs a lightweight sparse U-Net to predict the final fine-grained semantic occupancy within the completed scene.
    }
    \label{fig:sparseocc_plus}
    \vspace{-2mm}
\end{figure*}
\section{SparseOcc++}
To further exploit the potential of sparse representation, we present SparseOcc++, which features a novel \textit{geometry-aware sparse representation}, as defined in Sec.~\ref{sec:geo_aware_representation}. 
Fig.~\ref{fig:sparseocc_plus} presents an overview of the proposed SparseOcc++. 
It first extracts 2D features from images and dynamically initializes a set of sparse anchor voxels (Sec.~\ref{sec:anchor_vox}). 
To reconstruct the 3D geometry of occupied regions, these sparse anchor voxels function as queries to index corresponding image features via deformable cross attention (DCA), which are then used to complete the scene via scene completion field (SCF) learning (Sec.~\ref{sec:geo_reg}). Then, adapting to the predicted occupied regions, we use a Sparse U-net~\cite{unet} to extract sparse voxel features and predict the final occupancy results (Sec.~\ref{sec:sem_pred}).  

\subsection{Geometry-Aware Sparse Representation}
\label{sec:geo_aware_representation}
In prior approaches, scene completion is inherently entangled with high-dimensional semantic reasoning, leading to computational redundancy and geometric ambiguity (as discussed in Sec.~\ref{sec:joint_rep}). To address this limitation, we propose a novel \textit{geometry-aware sparse representation}. Our core strategy is to explicitly disentangle scene completion from semantic classification. Instead of achieving scene completion via indiscriminate semantic feature propagation and voxel-wise classification, we reformulate the problem as a regression task, allowing us to first focus on recovering the scene's structural extent before attending to semantics.

\noindent\textbf{Sparse Anchor Voxel.} To implement this regression-based scene completion, we draw inspiration from anchor-based 2D object detection methods such as Faster R-CNN~\cite{faster_rcnn}, which efficiently estimates object geometry via bounding-box regression from anchor boxes rather than dense pixel-wise classification. Analogously, we aim to recover the 3D scene geometry by regressing 3D geometric properties from \textit{sparse anchor voxels} $\mathbb{A}$. 
Unlike the sparse representation in Eq.~\ref{eq:sparse_representation}, which contains only coordinates and semantic embeddings, we augment each anchor with an explicit geometric attribute:
\begin{equation}
    \mathbb{A} = \{ (\mathbf{p}_i, \mathbf{f}_i, \mathbf{s}_i) \vert i=1,2,...N \},
\end{equation}
where $\mathbf{p}_i \in \mathbb{Z}^3$ denotes the coordinate, $\mathbf{f}_i \in \mathbb{R}^C$ is the learnable feature embedding, and $\mathbf{s}_i$ is the newly introduced geometric attribute designed to encode the scene structure.

\noindent\textbf{Scene Completion Field.} However, unlike object detection, which merely approximates an object's coarse envelope via bounding boxes, occupancy prediction demands the recovery of precise geometric boundaries. To define a geometric attribute $\mathbf{s}$ that captures this fine-grained structural detail, we leverage the theory of Signed Distance Fields (SDF), which is widely used in computer graphics. An SDF represents geometry as a continuous field, where the value indicates the distance to the nearest surface. Building on this theoretical foundation, we define a specialized \textit{scene completion field} (SCF). 
Different from standard SDFs that map coordinates to a single isotropic scalar representing the radius to the nearest surface point, our SCF is tailored to complete the complex, anisotropic shapes of outdoor objects.
Therefore, we formulate the SCF as a vector-valued function $\Phi$ conditioned on both spatial location $\mathbf{p}_i$ and semantic context $\mathbf{f}_i$:
\begin{equation}
\label{equ:scf}
\mathbf{s}_i = \Phi(\mathbf{p}_i, \mathbf{f}i) \in \mathbb{R}^K.
\end{equation}
Here, the geometric attribute $\mathbf{s}_i \in \mathbb{R}^K$ is a vector where each component $s$ represents the signed distance along a specific direction (e.g., orthogonal axes):
\begin{equation}
\label{equ:sdf}
    {s} = \begin{cases} 
      +\min_{\mathbf{v} \in \partial \Omega} \| \mathbf{p}_i - \mathbf{v} \|_2 & \text{if } \mathbb{A}_i \in \Omega_\text{empty} \\ 
      0 & \text{if } \mathbb{A}_i \in \partial \Omega \\
      -\min_{\mathbf{v} \in \partial \Omega} \| \mathbf{p}_i - \mathbf{v} \|_2 & \text{if } \mathbb{A}_i \in \Omega_\text{occupied} 
   \end{cases}
\end{equation}
where $\Omega \subset \mathbb{Z}^3$ denotes the voxel space and $\partial \Omega$ represents the geometric boundary along the corresponding direction of $s$.
Specifically, the sign of these distance components serves as a decisive indicator of occupancy status: positive values indicate empty space, while negative values signify the interiors of occupied regions.

By explicitly incorporating the SCF distance $\mathbf{s}_i$ into the sparse representation, we fundamentally shift the scene completion paradigm from exhaustive voxel-wise classification to {spatial proximity learning}. 
Consequently, we can efficiently determine the spatial extent of the scene, ensuring that subsequent fine-grained semantic predictions are strictly confined to the geometrically verified occupied voxels, thereby significantly reducing computational redundancy.

\noindent\textbf{Remarks: Connection and Distinction with Standard SDF.}
While our SCF is theoretically rooted in the Signed Distance Field, it differs fundamentally in three aspects tailored for occupancy prediction:
\begin{itemize}
\item \textit{Conditioning Context:} Standard SDFs typically map spatial coordinates $\mathbf{p}$ to distances for a static shape. In contrast, our SCF is a \textit{conditional field} $\Phi(\mathbf{p}, \mathbf{f})$ inferred from visual observations. It relies on semantic image features $\mathbf{f}$ to predict the missing geometries of occluded regions rather than merely fitting the observed surface voxels.
\item \textit{Anisotropy:} A standard SDF is a scalar field representing the isotropic distance to the nearest surface. Our SCF is formulated as a \textit{vector-valued field} in Eq.~\ref{equ:scf}. This vector formulation enables the encoding of anisotropic geometric information (e.g., decomposing into orthogonal components in Sec.~\ref{sec:geo_reg}), which is critical for completing the complex aspect ratios of outdoor objects.
\item \textit{Task Objective:} SDFs are often used for surface reconstruction or rendering. Our SCF is specifically designed for \textit{scene completion}, where the zero-level set serves as a geometric hull to constrain and propagate semantic information.
\end{itemize}

\subsection{Sparse Anchor Voxel Generation}
\label{sec:anchor_vox}
To efficiently recover the volumetric structure using the proposed SCF without exhaustively processing numerous empty candidate voxels, we introduce the concept of \textit{sparse anchor voxels}. 
Similar to anchor boxes in object detection~\cite{faster_rcnn,lidarrcnn}, these anchor voxels serve as initial geometric proposals from which our SCF is sampled and learned.

A straightforward implementation of sparse anchor voxel generation is to sample coordinates uniformly across the entire space. However, given the inherent sparsity of 3D scenes, in which occupied regions constitute only a small fraction of the total volume, uniform sampling incurs significant computational costs in empty spaces. To mitigate this, following~\cite{liu2023fully, lu2023octreeocc, wang2025opus}, we adopt a coarse-to-fine strategy to progressively generate anchor voxels strictly within geometrically important regions.

\noindent\textbf{Contextualization via Deformable Cross Attention.}
To determine the importance of each voxel, we must first endow it with image context. Given a set of sparse anchor voxel candidates $\mathbb{\tilde A}$, we utilize Deformable Cross Attention (DCA)~\cite{deformable_DETR} to aggregate multi-view image features.
Specifically, for each anchor candidate $(\mathbf{p}, \mathbf{f}) \in \mathbb{\tilde A}$, we project its 3D center $\mathbf{p}$ into the 2D image planes via the LiDAR-to-image transformation matrices $\mathcal{T}$, obtaining reference points $\pi(\mathbf{p}, \mathcal{T})$. These points guide the sampling of image features $\mathbf{F}^\text{2D}$. The updated feature is computed as:
\begin{equation}
    \text{DCA}(\mathbb{\tilde A}, \mathbf{F}^\text{2D}) = \frac{1}{N_\text{view}} \sum_{v=1}^{N_\text{view}} \sum_{j=1}^{N_s} w_{vj} \cdot \mathbf{F}^\text{2D}_v(\pi(\mathbf{p}, \mathcal{T}_v) + \Delta \mathbf{p}_{vj}),    
\end{equation}
where $N_s$ is the number of sampling points, $\Delta \mathbf{p}_{vj}$ is the learned sampling offset, and $w_{vj}$ is the attention weight. This mechanism aligns the 3D sparse anchors with their 2D visual evidence, thereby facilitating the effective selection of sparse anchor voxels.

\noindent\textbf{Pyramidal Generation Strategy.}
We instantiate the anchor generation as a multi-stage process across resolution levels $r \in \{0, 1, 2\}$, corresponding to $\{1/8, 1/4, 1/2\}$ of the original resolution.
Starting from an initial set $\mathbb{\tilde A}^{0}$ comprising dense voxels at the coarsest level (initialized with random embeddings), we iteratively refine the set to obtain $\mathbb{\tilde A}^{r}$:
\begin{equation}
    \begin{split}
    &\mathbb{S}^{r} = \text{Split}(\mathbb{\tilde A}^{r-1}), \\
    &\mathbb{\tilde A}^r = \text{Top-k} \left( \mathbb{S}^{r}, \mathcal{M}(\text{DCA}(\mathbb{S}^{r})), N^{r} \right).
    \end{split}
\end{equation}
Here, $\text{Split}(\cdot)$ denotes the geometric subdivision, where each parent voxel at level $r-1$ is split into 8 sub-voxels at level $r$. $\mathcal{M}(\cdot)$ is a lightweight MLP that predicts an occupancy score from DCA-aggregated features. $\text{Top-k}(\cdot)$ selects the top $N^r$ voxels with the highest scores to form the sparse anchor set for the current level.
In our implementation, we retain the top $\{1/5, 1/10\}$ candidates for levels 1 and 2, respectively.
Crucially, this cascade ensures that computation is concentrated solely on regions with high geometric saliency. The final set $\mathbb{\tilde A}^{2}$ is then utilized as the sparse anchor voxels $\mathbb{A}$ for the SCF-based scene completion described in Sec.~\ref{sec:geo_reg}.

\subsection{Compact Scene Completion}
\label{sec:geo_reg}
Building on the generated sparse anchors, this section details the learning of the scene completion field (SCF) $\Phi$. To accurately model complex outdoor scenes, we propose a specific orthogonal instantiation of the geometric vector $\mathbf{s}$, trained via a robust discretized learning strategy.

\begin{figure*}
    \centering
    \includegraphics[width=\linewidth]{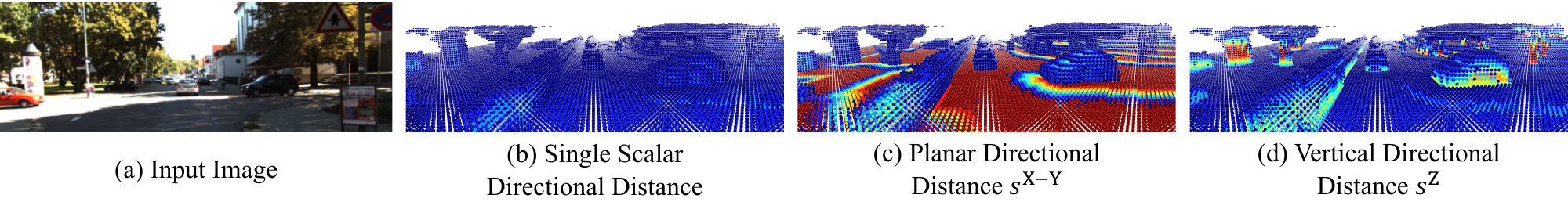}
    \caption{\textbf{Visualizations on directional distance.} (a) The input image. (b) A single scalar directional distance is insufficient to model the scene due to geometric anisotropy. (c)\&(d) The proposed orthogonal decomposition strategy decouples the directional distance into an X–Y planar component and a vertical Z-axis component, providing a closer match to the geometric distribution.}
    \label{fig:sdf_decoup}
    \vspace{-2mm}
\end{figure*}
\label{sec:dis_decomp}
\noindent\textbf{Orthogonal Decomposition Strategy.}
While the SCF definition in Eq.~\ref{equ:scf} allows for arbitrary directional distances, a naive isotropic design (i.e., treating $\mathbf{s}$ as a single scalar radius) is suboptimal for outdoor driving scenes due to strong \textit{geometric anisotropy}. For instance, `road' surfaces exhibit extensive horizontal spans but are extremely thin vertically, whereas `poles' are the inverse. As a result, a single scalar cannot simultaneously capture these conflicting spatial scales. As shown in Fig.~\ref{fig:sdf_decoup}(b), the SCF distance $\mathbf{{s}}$ is too small for sparse anchor voxels to model the scene.

To address this, we leverage the vector nature of our SCF and design $\mathbf{s}$ to explicitly decouple the scene along orthogonal axes. 
Specifically, we instantiate the SCF vector with two orthogonal components: the planar SCF distance ${s}^{\text{X-Y}}$ and the vertical SCF distance ${s}^{\text{Z}}$, i.e., $\mathbf{s} = ({s}^{\text{X-Y}}, {s}^{\text{Z}})$.
Specifically, for the anchor voxel at $\mathbf{p}_i=(x_i, y_i, z_i)$. The planar component ${s}^{\text{X-Y}}$ measures the distance to the boundary within the horizontal cross-section plane $\mathcal{P}_i = \{ (x, y, z) \mid z = z_i \}$, capturing the horizontal spread. 
Conversely, the vertical component ${s}^{\text{Z}}$ is constrained to the vertical axis $\mathcal{L}_i = \{ (x, y, z) \mid x = x_i, y = y_i \}$, capturing height information.
As shown in Fig.~\ref{fig:sdf_decoup}(c-d), this orthogonal decomposition enables the anchors to flexibly adapt to irregular shapes, significantly improving reconstruction accuracy.

\noindent\textbf{Discretized Learning.} Direct regression of continuous SCF values often suffers from convergence instability due to the infinite solution space and high variance inherent in outdoor environments. 
Inspired by recent advances in depth estimation~\cite{bevdepth}, we reformulate this regression problem as a \textit{classification} task.
Taking the planar SCF as an instance, we define a range $[-{s}^{\text{X-Y}}_{\text{max}}, {s}^{\text{X-Y}}_{\text{max}}]$ and discretize it into a set of bins $\mathbb{D} = \{d_1, d_2, \dots, d_{|\mathbb{D}|}\}$, where ${s}^{\text{X-Y}}_{\text{max}}$ is the maximum SCF planar distance.
Consequently, the regression task is transformed into a $|\mathbb{D}|$-way classification problem, where $|\mathbb{D}| = 2{s}^{\text{X-Y}}_{\text{max}} + 1$.
For each anchor voxel, we employ an MLP head to predict the probability distribution $\mathbf{P}^{\text{X-Y}}$ over $\mathbb{D}$. The final distance is determined by the class index with the maximum likelihood:
\begin{equation} 
{\hat{s}}^{\text{X-Y}}_i = d_{k} \quad \text{where} \quad k = \operatorname*{argmax}_{k \in {1, \dots, \|\mathbb{D}\|}} \mathbf{P}^{\text{X-Y}}_i(k). 
\end{equation}

Similarly, the vertical component ${s}^{\text{Z}}$ is learned using the same formulation. This {discretized geometry learning} strategy stabilizes training convergence and improves robustness to sensor noise.

\noindent\textbf{Geometry-Guided Propagation.}
\begin{algorithm}[t]
\caption{PyTorch-like pseudocode of geometry-guided propagation.}
\label{alg:propagation}
\definecolor{codeblue}{rgb}{0.25,0.5,0.5}
\definecolor{codegreen}{rgb}{0.0,0.6,0.0}
\lstset{
  backgroundcolor=\color{white},
  basicstyle=\fontsize{7.5pt}{8.5pt}\ttfamily\selectfont,
  columns=fullflexible,
  breaklines=true,
  captionpos=b,
  commentstyle=\color{codegreen},
  keywordstyle=\color{blue},
}
\begin{lstlisting}[language=python]
# s_pred: predicted SCF distance (N)
# p_anchor: coordinates of anchor voxels (N, 3)
# f_anchor: features of anchor voxels (N, C)

# Initialize binary occupancy map
M = torch.zeros(H, W, D)
M[p_anchor] = 1

for s_j in range(1, s_max + 1): 
    # Anchors predicted to have distance -s_j
    M_j = (s_pred == -s_j)
    p_j = p_anchor[M_j]

    if len(p_j) == 0: continue
    
    M_j = torch.zeros_like(M); M_j[p_j] = 1
    # Propagate occupancy from anchors to local neighbors within radius s
    M_prop = F.max_pool3d(M_j, kernel_size=2*s_j+1, stride=1, padding=s)
    M += M_prop # Accumulate

# Extract all occupied voxel coordinates after padding
p_total = torch.where(M > 0)
p_new_diff = torch.tensor(set(p_total) - set(p_anchor))
num_new = p_total.shape[0] - p_anchor.shape[0]

# New voxels are zero-initialized
f_new = torch.zeros(num_new, C)  # Zero-init
p_final = torch.cat([p_anchor, p_new_diff], dim=0)
f_final = torch.cat([f_anchor, f_new], dim=0)

return p_final, f_final
\end{lstlisting}
\end{algorithm}
Once the sparse anchor voxels with their predicted SCF distances are obtained, we proceed to complete the scene.
The predicted SCF distance ${\hat{s}}$ serves as a strong geometric prior, indicating the spatial support of the object centered at each anchor. To complete the scene while maintaining efficiency, we introduce a \textit{Geometry-Guided Propagation} strategy (Algorithm~\ref{alg:propagation}). Instead of indiscriminately densifying the space using the sparse latent propagator, we propagate the occupancy status from each sparse anchor to its local neighborhood, strictly guided by the predicted distance ${\hat{s}}$. 
Specifically, we treat ${\hat{s}}$ as a propagation radius. By iterating through quantized distance values ${\hat{s}}_j$, we apply a 3D Max-Pooling operation with a kernel size of $2{\hat{s}}_j + 1$ on the corresponding anchors. This operation effectively expands the single occupied anchor voxel into a local block with a radius of ${\hat{s}}_j$, completing the missing voxels inside the scene boundary.
This strategy ensures that scene completion is geometry-aware: new voxels are generated only where the SCF predicts the existence of an object, thereby avoiding computational waste from processing empty regions.

\noindent\textbf{SCF Ground Truth Generation.}
\begin{algorithm}[t]
\caption{PyTorch-like pseudocode for SCF GT generation.}
\label{alg:sdf_gt}
\definecolor{codeblue}{rgb}{0.25,0.5,0.5}
\definecolor{codegreen}{rgb}{0.0,0.6,0.0}
\lstset{
  backgroundcolor=\color{white},
  basicstyle=\fontsize{7.5pt}{8.5pt}\ttfamily\selectfont,
  columns=fullflexible,
  breaklines=true,
  captionpos=b,
  commentstyle=\color{codegreen},
  keywordstyle=\color{blue},
}
\begin{lstlisting}[language=python]
# occ_gt: occupancy ground-truth (H, W, D)
# s_max: predefined maximum SCF distance value

# Initialize binary occupancy gt and SCF gt
occ_gt[occ_gt > 0] = 1
scf_gt = torch.zeros_like(occ_gt)

for s_j in range(1, s_max + 1): 
    # Define kernel size covering the area of radius s_j
    # Note: Kernel dimensions vary for Planar vs Vertical SCF GT generation.
    k_j = 2 * s_j + 1  
    
    pool_val = F.avg_pool3d(occ_gt.float(), kernel_size=k_j, stride=1, padding=k_j // 2)
    
    # If a window is pure 1s, avg is 1. If pure 0s, avg is 0.
    # So if pool_val == occ_gt, the radius s_j is valid.
    is_homogeneous = (pool_val == occ_gt)

    # Update distance for occupied voxels (Negative)
    mask_inner = is_homogeneous & (occ_gt == 1)
    scf_gt[mask_inner] = -s_j

    # Update distance for empty voxels (Positive)
    mask_outer = is_homogeneous & (occ_gt == 0)
    scf_gt[mask_outer] = s_j

return scf_gt
\end{lstlisting}
\end{algorithm}
To supervise SCF learning, we efficiently generate ground truth (GT) from semantic occupancy labels. 
The process is outlined in Algorithm~\ref{alg:sdf_gt}. First, we simplify the semantic labels into a binary geometric occupancy grid $\mathbf{G}$ (1 for occupied, 0 for empty). 
Instead of using an expensive Euclidean Distance Transform (EDT), we employ an iterative 3D average-pooling strategy. 
For a candidate distance ${s}_j$, we apply 3D average pooling with a kernel size $K = 2{s}_j+1$ on $\mathbf{G}$. If a voxel's value remains unchanged after pooling, it indicates that its neighborhood within a radius ${s}_j$ is homogeneous (purely occupied or empty), confirming that its distance is at least ${s}_j$. By iterating ${s}_j$ from 1 to ${s}_{\text{max}}$, we resolve the distance values for all voxels.
Consistent with our decomposition strategy, we generate GT separately for planar distance ${s}^{\text{X-Y}}$ and vertical distance ${s}^{\text{Z}}$ by adjusting the kernel dimensions: $K^{\text{X-Y}}=[2{s}^{\text{X-Y}}+1, 2{s}^{\text{X-Y}}+1, 1]$ for the horizontal plane and $K^{\text{Z}}=[1, 1, 2{s}^{\text{Z}}+1]$ for the vertical axis.


\subsection{Geometry-Constrained Semantic Prediction}
\label{sec:sem_pred}
Upon scene completion, we obtain a refined voxel set $\mathbb{V}_{\text{occ}}$. However, a feature gap exists: while the original anchor voxels possess learned embeddings, the newly completed voxels are initially featureless (zero-initialized).
To bridge this gap, we reapply Deformable Cross Attention (DCA) to the complete voxel set $\mathbb{V}_{\text{occ}}$. This enables newly generated voxels to query their projected locations in multi-view images and to aggregate discriminative visual features.
Finally, these geometrically refined and semantically enriched sparse voxels are processed by a lightweight Sparse U-Net~\cite{unet}, which facilitates local context exchange among neighboring voxels, yielding the final semantic occupancy results $\mathbb{\hat{Y}}$.

\subsection{Objective Function}
Following the previous occupancy prediction method~\cite{OpenOccupancy, monoscene}, we employ cross-entropy loss $\mathcal{L}_\mathrm{ce}^\mathrm{occ}$, Lovasz loss $\mathcal{L}_\mathrm{lovasz}$, affinity loss $\mathcal{L}_\mathrm{scale}^\mathrm{geo}$, and $\mathcal{L}_\mathrm{scale}^\mathrm{sem}$ to supervise the final occupancy $\mathbb{Y}$:

\begin{equation}
\label{equ:loss_occ}
\mathcal{L}_\mathrm{occ} =  \mathcal{L}_\mathrm{ce}^\mathrm{occ}+\mathcal{L}_\mathrm{lovasz}+\mathcal{L}_\mathrm{scale}^\mathrm{geo}+\mathcal{L}_\mathrm{scale}^\mathrm{sem}.    
\end{equation}

We also apply another loss $\mathcal{L}_\mathrm{anchor}$, which is the same occupancy loss in Eq.~\ref{equ:loss_occ}, to supervise the multi-level predictions on $\mathbb{\tilde A}$ and guide the sparse anchor voxel generation in Sec.~\ref{sec:anchor_vox}. Besides, the learning of SCF in Sec.~\ref{sec:geo_reg} is also supervised by a cross-entropy loss $\mathcal{L}_\mathrm{scf} = \mathcal{L}_\mathrm{ce}^\mathrm{scf}$ as a classification task.

Hence, the overall objective function is formulated as:
\begin{equation}
\mathcal{L} =  \mathcal{L}_\mathrm{occ}+\mathcal{L}_\mathrm{anchor}+\mathcal{L}_\mathrm{dist}.
\end{equation}
\section{Experiments}
This section provides a comprehensive evaluation of SparseOcc and SparseOcc++. 

\begin{table*}[ht]
    \footnotesize
    \setlength{\tabcolsep}{0.0035\linewidth}
    \newcommand{\classfreq}[1]{{~\tiny(\semkitfreq{#1}\%)}}  %
    \centering
    \caption{\textbf{Vision-based occupancy prediction results on the SemanticKITTI~\cite{semantickitti} validation set.} For the accuracy evaluation, we report the geometric metric (IoU), the semantic metric (mIoU), and the IoU for each individual semantic class. For the efficiency evaluation, we report the frames per second (FPS). The \textbf{bold} numbers indicate the best results. The methods marked by ``*" are reported by~\cite{monoscene}. }
    \resizebox{1\linewidth}{!}{
    \begin{tabular}{l|c c| c c c c c c c c c c c c c c c c c c c|c}
        \toprule
        Method & IoU & mIoU
        & \rotatebox{90}{\textcolor{road}{$\blacksquare$} road\classfreq{road}} 
        & \rotatebox{90}{\textcolor{sidewalk}{$\blacksquare$} sidewalk\classfreq{sidewalk}}
        & \rotatebox{90}{\textcolor{parking}{$\blacksquare$} parking\classfreq{parking}} 
        & \rotatebox{90}{\textcolor{other-ground}{$\blacksquare$} other-ground\classfreq{otherground}} 
        & \rotatebox{90}{\textcolor{building}{$\blacksquare$} building\classfreq{building}} 
        & \rotatebox{90}{\textcolor{car}{$\blacksquare$} car\classfreq{car}} 
        & \rotatebox{90}{\textcolor{truck}{$\blacksquare$} truck\classfreq{truck}} 
        & \rotatebox{90}{\textcolor{bicycle}{$\blacksquare$} bicycle\classfreq{bicycle}} 
        & \rotatebox{90}{\textcolor{motorcycle}{$\blacksquare$} motorcycle\classfreq{motorcycle}} 
        & \rotatebox{90}{\textcolor{other-vehicle}{$\blacksquare$} other-vehicle\classfreq{othervehicle}} 
        & \rotatebox{90}{\textcolor{vegetation}{$\blacksquare$} vegetation\classfreq{vegetation}} 
        & \rotatebox{90}{\textcolor{trunk}{$\blacksquare$} trunk\classfreq{trunk}} 
        & \rotatebox{90}{\textcolor{terrain}{$\blacksquare$} terrain\classfreq{terrain}} 
        & \rotatebox{90}{\textcolor{person}{$\blacksquare$} person\classfreq{person}} 
        & \rotatebox{90}{\textcolor{bicyclist}{$\blacksquare$} bicyclist\classfreq{bicyclist}} 
        & \rotatebox{90}{\textcolor{motorcyclist}{$\blacksquare$} motorcyclist\classfreq{motorcyclist}} 
        & \rotatebox{90}{\textcolor{fence}{$\blacksquare$} fence\classfreq{fence}} 
        & \rotatebox{90}{\textcolor{pole}{$\blacksquare$} pole\classfreq{pole}} 
        & \rotatebox{90}{\textcolor{traffic-sign}{$\blacksquare$} traffic-sign\classfreq{trafficsign}} 
        & FPS \\
        \midrule
        \multicolumn{23}{c}{\textit{Dense Occupancy Methods}} \\
        \midrule
        LMSCNet*~\cite{roldao2020lmscnet}                   & 28.61             & 6.70              & 40.68             & 18.22             & 4.38              & 0.00              & 10.31 & 18.33             & 0.00          & 0.00 & 0.00 & 0.00 & 13.66 & 0.02 & 20.54 & 0.00 & 0.00 & 0.00 & 1.21 & 0.00 & 0.00 & -  \\ %
        3DSketch*~\cite{3d-sketch}                          & 33.30             & 7.50              & 41.32             & 21.63             & 0.00              & 0.00              & 14.81 & 18.59             & 0.00          & 0.00 & 0.00 & 0.00 & 19.09 & 0.00 & 26.40 & 0.00 & 0.00 & 0.00 & 0.73 & 0.00 & 0.00 & - \\ %
        AICNet*~\cite{li2020anisotropic}                    & 29.59             & 8.31              & 43.55             & 20.55             & 11.97             & 0.07              & 12.94 & 14.71             & 4.53          & 0.00 & 0.00 & 0.00 & 15.37 & 2.90 & 28.71 & 0.00 & 0.00 & 0.00 & 2.52 & 0.06 & 0.00 & -  \\ %
        JS3C-Net*~\cite{js3cnet}                            & 38.98    & 10.31             & 50.49             & 23.74             & 11.94             & 0.07              & 15.03 & 24.65             & 4.41          & 0.00 & 0.00 & 6.15 & 18.11 & 4.33 & 26.86 & 0.67 & 0.27 & 0.00 & 3.94 & 3.77 & 1.45  & -\\
        MonoScene~\cite{monoscene}                 & 36.86             & 11.08             & 56.52             & 26.72             & 14.27             & 0.46              & 14.09 & 23.26             & 6.98          & 0.61 & 0.45 & 1.48 & 17.89 & 2.81 & 29.64 & 1.86 & 1.20 & 0.00 & 5.84 & 4.14 & 2.25 & 3.9\\
        TPVFormer~\cite{tpvformer}                          & 35.61             & 11.36             & 56.50             & 25.87             & 20.60    & 0.85     & 13.88 & 23.81             & 8.08          & 0.36 & 0.05 & 4.35 & 16.92 & 2.26 & 30.38 & 0.51 & 0.89 & 0.00 & 5.94 & 3.14 & 1.52 & -\\
        NDC-Scene~\cite{yao2023ndc}                         & 37.24             & 12.70             & 59.20    & 28.24    & 21.42    & \textbf{1.67}     & 14.94 & 26.26             & 14.75         & 1.67 & 2.37          & 7.73           & 19.09          & 3.51           & 31.04          & 3.60          & 2.74          & 0.00          & 6.65          & 4.53              & 2.73     & 4.5     \\
        OccFormer~\cite{occformer}                          & 36.50             & 13.46    & 58.85             & 26.88             & 19.61             & 0.31              & 14.40 & 25.09    & \textbf{25.53} & 0.81 & 1.19 & 8.52 & 19.63 & 3.93 & \textbf{32.62} & 2.78 & 2.82 & 0.00 & 5.61 & 4.26 & 2.86 & 4.2 \\
        \midrule
        \multicolumn{23}{c}{\textit{Two-Stage Methods}} \\
        \midrule
        VoxFormer~\cite{voxformer}           & \textbf{44.02} & 12.35          & 54.76          & 26.35          & 15.50          & 0.70           & 17.65          & 25.79          & 5.63           & 0.59          & 0.51          & 3.77           & 24.39          & 5.08           & 29.96          & 1.78          & 3.32 & 0.00          & 7.64          & 7.11          & 4.18 & 6.0          \\
        Symphonies~\cite{jiang2024symphonize}               & 41.92             & \textbf{14.89}    & 56.37          & 27.58          & 15.28          & 0.95           & \textbf{21.64} & \textbf{28.68} & 20.44          & \textbf{2.54} & 2.82 & \textbf{13.89} & \textbf{25.72} & \textbf{6.60} & 30.87           & 3.52 & 2.24          & 0.00          & \textbf{8.40} & \textbf{9.57} & \textbf{5.76} & 5.2 \\
        \midrule
        \multicolumn{23}{c}{\textit{Sparse Occupancy Methods}} \\
        \midrule
        SparseOcc~(ours)                  & 36.48             & 13.12             & \textbf{59.59}    & \textbf{29.68}    & 20.44             & 0.47              & 15.41    & 24.03         & 18.07 & 0.78 & 0.89 & 8.94 & 18.89 & 3.46 & 31.06 & 3.68 & 0.62 & 0.00 & 6.73 & 3.89 & 2.60 & 6.7 \\

        
        SparseOcc++~(ours)                       & 37.08            & 13.23             & 58.46    & 28.73    & 19.85             & 1.35              & 14.74    & 26.26         & 20.34 & 0.82 & 2.13 & 8.07 & 18.11 & 2.93 & 30.75 & 3.60 & 1.58 & 0.00 & 6.66 & 4.27 & 2.74 & \textbf{24.7} \\
        
        $\text{SparseOcc++}_{\text{swin}}$~(ours)  & 37.32             & 13.97            & 59.08    & 29.43    & \textbf{22.06}             & 0.50              & 15.29    & 25.68         & 18.45 & 1.34 & \textbf{3.07} & 12.72 & 19.34 & 3.28 & 31.82 & \textbf{4.17} & \textbf{4.63} & 0.00 & 7.48 & 4.29 & 2.75 & 20.1 \\
        \hline
    \end{tabular}}\\
    \label{table:kitti}
    \vspace{-2mm}
\end{table*}
\subsection{Datasets}
\label{sec:datasets}
We evaluate our proposed sparse representations on two popular benchmarks: SemanticKITTI~\cite{semantickitti} and nuScenes-Occupancy~\cite{OpenOccupancy}.

\noindent\textbf{SemanticKITTI~\cite{semantickitti}} consists of 22 sequences comprising monocular images, LiDAR point clouds, LiDAR segmentation labels, and semantic scene completion labels. The dataset split is defined as follows: sequence 08 is reserved for validation, sequences 00-10 (excluding 08) are used for training, and sequences 11-21 are set aside for testing. Each 3D voxel is annotated as either empty or one of 19 semantic classes.

\noindent\textbf{nuScenes-Occupancy~\cite{OpenOccupancy}} extends the widely adopted nuScenes~\cite{nuscenes} dataset by providing dense 3D semantic occupancy annotations for keyframes. These annotations, which include one empty class and 16 semantic classes, were generated using the Augmenting and Purifying pipeline. The dataset comprises 700 training scenes and 150 validation scenes from nuScenes. The original nuScenes dataset provides the corresponding multi-view images and LiDAR point clouds.

\subsection{Evaluation Metrics}
\label{sec:eval_metrics}
Following standard practice~\cite{monoscene, huang2024gaussianformer,occformer}, we employ Intersection-over-Union (IoU) to evaluate class-agnostic geometry completion and mean IoU (mIoU) across all semantic classes as the semantic metric:
\begin{equation}
    \text{IoU} = \frac{\text{TP}_{\neq 0}}{\text{TP}_{\neq 0} + \text{FP}_{\neq 0} + \text{FN}_{\neq 0}}, 
\end{equation}
\begin{equation}
    \text{mIoU} = \frac{1}{N_\text{cls}} \sum_{i=1}^{N_\text{cls}} \frac{\text{TP}_i}{\text{TP}_i + \text{FP}_i + \text{FN}_i}, \quad 
\end{equation}
where $N_\text{cls}$ denotes the largest index of non-empty classes, 0 represents the empty class, and $ \text{TP} $, $ \text{FP} $, and $ \text{FN} $ denote the counts of true positives, false positives, and false negatives, respectively.

\subsection{Implementation Details}
\label{sec:impl_details}
\noindent\textbf{Input \& Backbone.} 
We set the input image resolutions to $(384, 1280)$ for SemanticKITTI and $(256, 704)$ for nuScenes-Occupancy~\cite{OpenOccupancy}. For the image encoder, to ensure a fair comparison with prior work~\cite {monoscene,voxformer, OpenOccupancy}, we primarily use ResNet-50~\cite{resnet} in most experiments. Specifically for SemanticKITTI, we follow~\cite{occformer} by employing EfficientNetB7~\cite{efficientnet} with SecondFPN. Additionally, to demonstrate the scalability and robustness of our method, we introduce a variant named $\text{SparseOcc++}_{\text{swin}}$ equipped with a stronger Swin Transformer backbone~\cite{swin_transformer}.

\noindent\textbf{Network Architecture Details.} 
For SparseOcc, the 2D-to-3D view transformation is implemented following~\cite{occformer,OpenOccupancy}. It generates 3D feature volumes of size 128$\times$128$\times$16 and 128$\times$128$\times$10, with 128 channels for SemanticKITTI and nuScenes-Occupancy, respectively, before converting them into a sparse representation.. We stack the sparse propagator for $L=4$ times, with each layer followed by a sparse convolution (stride 2) for downsampling. The kernel size $k$ for the orthogonal convolutions in the sparse completion and contextual aggregation blocks is set to 3. The sparse voxel decoder projects the multi-scale features to 192 channels, enhancing them via interpolation and summation. The sparse transformer head consists of 9 layers for query updating and decoding, utilizing $N_q=100$ queries. For SparseOcc++, during compact scene completion (Sec.~\ref{sec:geo_reg}), we set the maximum planar SCF distance to $s_\text{max}^{\text{X-Y}}=10$ and the maximum vertical SCF distance to $s_\text{max}^{\text{Z}}=5$. This completion step is applied once to the final output of the sparse anchor voxel generation module.
The lightweight sparse U-Net~\cite{unet} comprises four layers, each consisting of one sparse residual block~\cite{zhang2024safdnet}.

\noindent\textbf{Training Strategy.}
The learning rate undergoes a warmup phase for the first 100 iterations on SemanticKITTI and 500 iterations on nuScenes-Occupancy, followed by a cosine decay schedule. All models are trained for 30 epochs with a batch size of 8. During training, we apply standard data augmentations, including random flipping, rotation, and resizing, to both input images and the 3D occupancy GT. Following~\cite{occformer,mask2former}, we sample 50,176 points based on class frequency for supervision to accelerate the training of SparseOcc.

\noindent\textbf{Inference \& Efficiency.}
During inference, the predicted masks are upsampled $2\times$ and $4\times$ via trilinear interpolation to match the ground-truth sizes of SemanticKITTI and nuScenes-Occupancy, respectively. For efficiency analysis, we report Frames Per Second (FPS) evaluated on a single NVIDIA A100 GPU with a batch size of 1 in FP32 precision. To ensure a fair comparison, no custom kernels are utilized beyond the standard deformable CUDA operator.

\begin{table*}[ht]
    \scriptsize
    \setlength{\tabcolsep}{3pt}
	\centering
    \caption{\textbf{Vision-based occpancy prediction results on the nuScenes-Occupancy~\cite{OpenOccupancy} validation set.} For the accuracy evaluation, we report the geometric metric (IoU), the semantic metric (mIoU), and the IoU for each individual semantic class. For the efficiency evaluation, we report the frames per second (FPS) and FLOPs. The \textbf{bold} numbers indicate the best results.}
    \resizebox{1\linewidth}{!}
    {
	\begin{tabular}{l|c c | c c c c c c c c c c c c c c c c | c c}
 
		\toprule
		Method
		& \makecell[c]{IoU}
            & \makecell[c]{mIoU}
		& \rotatebox{90}{\textcolor{barrier}{$\blacksquare$} barrier} 
		& \rotatebox{90}{\textcolor{bicycle}{$\blacksquare$} bicycle}
		& \rotatebox{90}{\textcolor{bus}{$\blacksquare$} bus} 
		& \rotatebox{90}{\textcolor{car}{$\blacksquare$} car} 
		& \rotatebox{90}{\textcolor{const. veh.}{$\blacksquare$} const. veh.} 
		& \rotatebox{90}{\textcolor{motorcycle}{$\blacksquare$} motorcycle} 
		& \rotatebox{90}{\textcolor{pedestrian}{$\blacksquare$} pedestrian} 
		& \rotatebox{90}{\textcolor{traffic cone}{$\blacksquare$} traffic cone} 
		& \rotatebox{90}{\textcolor{trailer}{$\blacksquare$} trailer} 
		& \rotatebox{90}{\textcolor{truck}{$\blacksquare$} truck} 
		& \rotatebox{90}{\textcolor{drive. suf.}{$\blacksquare$} drive. suf.} 
		& \rotatebox{90}{\textcolor{other flat}{$\blacksquare$} other flat} 
		& \rotatebox{90}{\textcolor{sidewalk}{$\blacksquare$} sidewalk} 
		& \rotatebox{90}{\textcolor{terrain}{$\blacksquare$} terrain} 
		& \rotatebox{90}{\textcolor{manmade}{$\blacksquare$} manmade} 
		& \rotatebox{90}{\textcolor{vegetation}{$\blacksquare$} vegetation}
        & \makecell[c]{FPS}
        & \makecell[c]{FLOPs}
            \\
		\midrule
        MonoScene~\cite{monoscene} & 18.4 & 6.9 & 7.1  & 3.9  &  9.3 &  7.2 & 5.6  & 3.0  &  5.9& 4.4& 4.9 & 4.2 & 14.9 & 6.3  & 7.9 & 7.4  & \textbf{10.0} & 7.6 & -\\
  
        TPVFormer~\cite{tpvformer} & 15.3 &  7.8 & 9.3  & 4.1  &  11.3 &  10.1 & 5.2  & 4.3  & 5.9 & 5.3&  6.8& 6.5 & 13.6 & 9.0  & 8.3 & 8.0  & 9.2 & 8.2 & 2.3 & 1132G \\
  
        OpenOccupancy~\cite{OpenOccupancy}  & 19.3  & 10.3  &  9.9 & 6.8  & 11.2  & 11.5  & 6.3  & 8.4  & 8.6 & 4.3 & 4.2 & 9.9 & 22.0  & 15.8 & 14.1  & 13.5  & 7.3 & 10.2  & 1.4 & 1716G \\
        
        C-CONet~\cite{OpenOccupancy} & 20.1  & 12.8&13.2  & 8.1 &  \textbf{15.4} &  17.2 & 6.3  & 11.2  & 10.0  &  8.3 & 4.7 & 12.1 & 31.4 & 18.8 & 18.7  & 16.3 & 4.8  &8.2 & 0.7 & 1810G \\

        C-OccGen~\cite{wang2024occgen} & 23.4 & \textbf{14.5} & 15.5 & 9.1 & 15.3 & \textbf{19.2} & 7.3 & \textbf{11.3} & \textbf{11.8} & 8.9 & 5.9 & \textbf{13.7} & \textbf{34.8} & \textbf{22.0} & \textbf{21.8} & \textbf{19.5} & 6.0 &  9.9 & - & - \\

        SparseOcc~(ours) & 21.8 & 14.1 & \textbf{16.1} & \textbf{9.3} & 15.1 & 18.6 & 7.3 & 9.4 & 11.2 & \textbf{9.4} & \textbf{7.2} & 13.0 & 31.8 & 21.7 & 20.7 & 18.8 & 6.1 & 10.6 & 4.0 & 455G \\
        
        SparseOcc++~(ours) & \textbf{24.1} & 14.1 & 14.5 & 8.1 & 14.7 & 18.1 & \textbf{7.6} & 9.4 & 11.0 & 7.1 & 6.6 & 13.3 & 32.9 & 21.4 & 21.4 & \textbf{19.5} & 8.0 & \textbf{12.6} & \textbf{19.7} & \textbf{183G}\\
        
	\bottomrule
	\end{tabular}}\\
	\label{table:nusc}
    \vspace{-2mm}
\end{table*}
\subsection{Benchmark Results}
\label{sec:benchmark_results}
\noindent\textbf{SemanticKITTI.} 
In Tab.~\ref{table:kitti}, we present a quantitative comparison of our SparseOcc and SparseOcc++ against existing monocular approaches on the SemanticKITTI validation set. 
Overall, our methods deliver comparable or superior performance. 
Notably, compared with the dense-occupancy baseline, we achieve higher IoU and mIoU. In addition, VoxFormer~\cite{voxformer} and Symphonies~\cite{jiang2024symphonize} adopt a two-stage occupancy framework: they first use pretrained models to estimate occupancy or depth maps, followed by a dense occupancy network to generate final predictions. While this yields strong performance, the FPS reported for these methods accounts only for the second-stage inference. In contrast, our improved variant, SparseOcc++, is fully end-to-end and considerably faster.
Furthermore, we observe that SparseOcc++ provides notably better results for small object categories such as `person', `bicyclist', and `motorcycle'. This indicates that our approach adaptively propagates features according to the relative positions of voxels with respect to the geometric surface. 
Crucially, our sparse representation offers distinct efficiency gains: SparseOcc++ operates at a real-time rate of 24.7 FPS, making it $5.9\times$ faster than OccFormer~\cite{occformer}.

\noindent\textbf{nuScenes-Occupancy.} Tab.~\ref{table:nusc} quantitatively compares the proposed sparse representations with several previous works. 
Specifically, our SparseOcc outperforms the 3D dense representation based C-CONet~\cite{OpenOccupancy} by 1.7\% in geometry IoU and 1.3\% in semantic mIoU. This result underscores the effectiveness of SparseOcc.
Besides, we observe that the model's FLOPs are reduced by 74.9\% relative to the baseline, further justifying the efficiency of our SparseOcc. 
Furthermore, benefiting from the proposed geometry-aware sparse representation, our SparseOcc++ achieves the highest geometric IoU, surpassing the dense occupancy network C-OccGen~\cite{wang2024occgen} by 0.7\% and the entangled geometry-semantic sparse method SparseOcc by 2.3\%. 
Note that the spatial resolution of nuScenes-OpenOccupancy is $512 \times 512 \times 32$, with each frame containing 6 surrounding images. 
Consequently, while the inference speed decreases to 19.7 FPS, it remains significantly faster than previous methods~\cite{tpvformer,OpenOccupancy}.

\begin{figure*}[t!] 
    \centering

    \newcolumntype{P}[1]{>{\centering\arraybackslash}m{#1}}
    \setlength{\tabcolsep}{0.001\textwidth}
    \begin{tabular}{P{1.\textwidth}}		
    \\[-0.8em]
    \includegraphics[width=1.\linewidth]{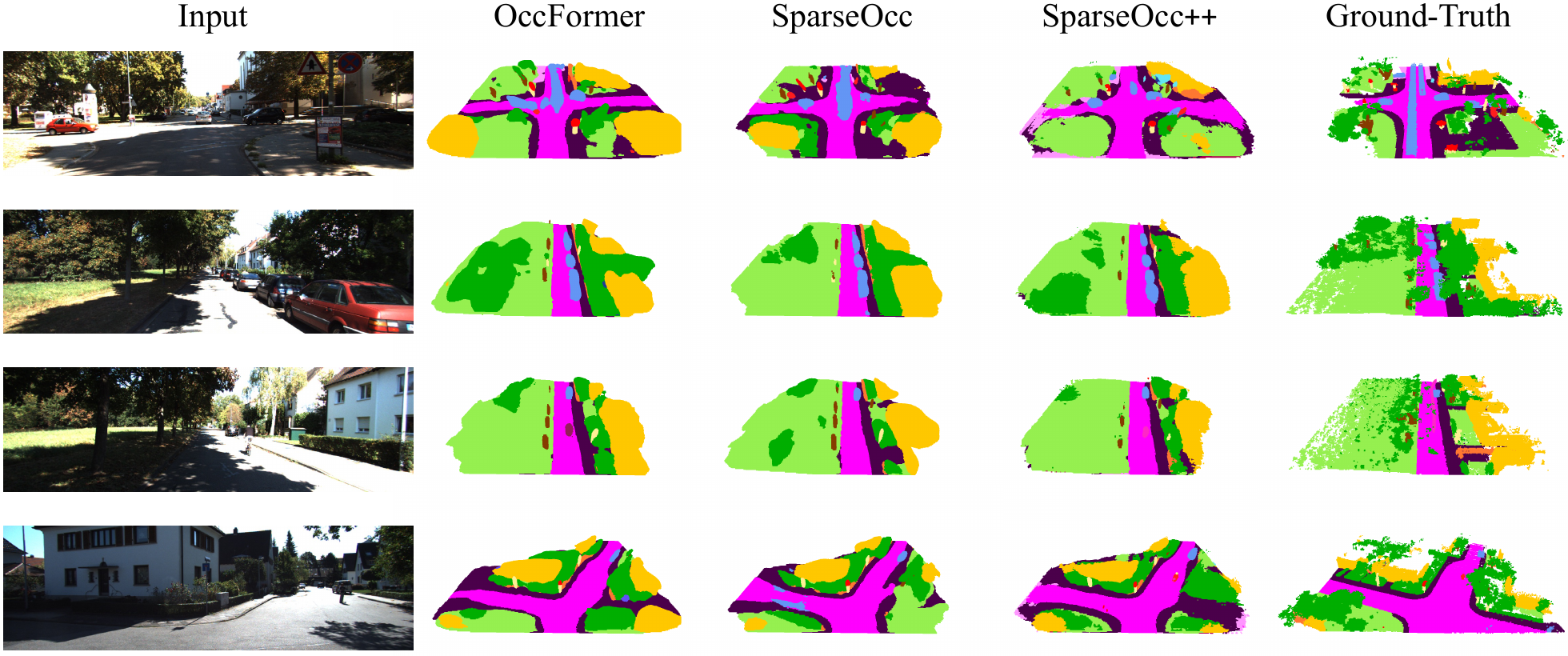}  \\
    \multicolumn{1}{c}{
    \textcolor{bicycle}{$\blacksquare$} bicycle~
    \textcolor{car}{$\blacksquare$} car~
    \textcolor{motorcycle}{$\blacksquare$} motorcycle~
    \textcolor{truck}{$\blacksquare$} truck~
    \textcolor{other-vehicle}{$\blacksquare$} other vehicle~
    \textcolor{person}{$\blacksquare$} person~
    \textcolor{bicyclist}{$\blacksquare$} bicyclist~
    \textcolor{motorcyclist}{$\blacksquare$} motorcyclist~
    \textcolor{road}{$\blacksquare$} road~
    \textcolor{parking}{$\blacksquare$} parking~}\\
    \multicolumn{1}{c}{
    \textcolor{sidewalk}{$\blacksquare$} sidewalk~
    \textcolor{other-ground}{$\blacksquare$} other ground~
    \textcolor{building}{$\blacksquare$} building~
    \textcolor{fence}{$\blacksquare$} fence~
    \textcolor{vegetation}{$\blacksquare$} vegetation~
    \textcolor{trunk}{$\blacksquare$} trunk~
    \textcolor{terrain}{$\blacksquare$} terrain~
    \textcolor{pole}{$\blacksquare$} pole~
    \textcolor{traffic-sign}{$\blacksquare$}traffic sign			
    }
    \end{tabular}
    \caption{\textbf{Qualitative results on the SemanticKITTI validation set.} The input monocular image is shown on the leftmost side, and the 3D occupancy predictions of the dense baseline OccFormer~\cite{occformer}, SparseOcc, SparseOcc++, and the ground truth are then visualized sequentially. The regions highlighted by rectangles show areas with noticeable differences. Best viewed in color and zoomed in.}
    \label{fig:vis_kitti}

    \vspace{5mm}
    
    \newcolumntype{P}[1]{>{\centering\arraybackslash}m{#1}}
    \setlength{\tabcolsep}{0.001\textwidth}
    \begin{tabular}{P{1.\textwidth}}		
    \\[-0.8em]
    \includegraphics[width=1.\linewidth]{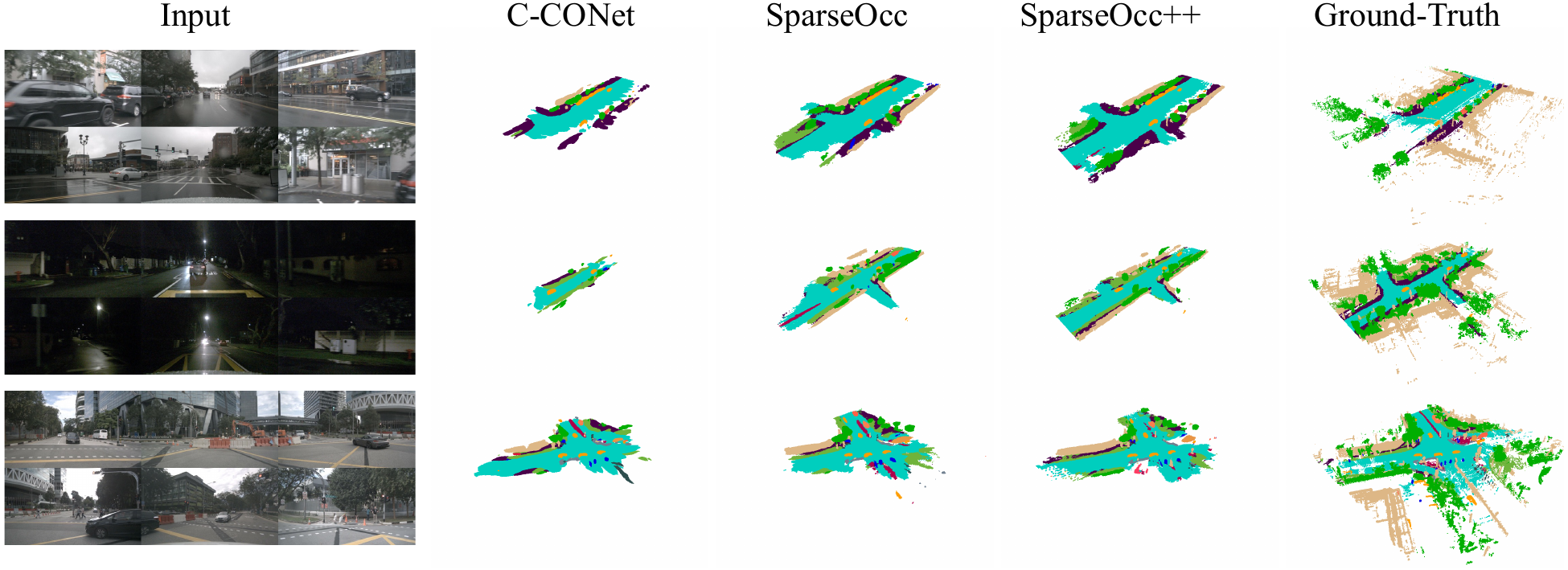}  \\
    \multicolumn{1}{c}{
    \textcolor{barrier}{$\blacksquare$} barrier~
    \textcolor{bicycle}{$\blacksquare$} bicycle~
	{\textcolor{bus}{$\blacksquare$} bus~} 
	{\textcolor{car}{$\blacksquare$} car~} 
	{\textcolor{const. veh.}{$\blacksquare$} const. veh.~} 
	{\textcolor{motorcycle}{$\blacksquare$} motorcycle~} 
	{\textcolor{pedestrian}{$\blacksquare$} pedestrian~} 
	{\textcolor{traffic cone}{$\blacksquare$} traffic cone~} 
	{\textcolor{trailer}{$\blacksquare$} trailer~} 
	{\textcolor{truck}{$\blacksquare$} truck~}} \\
    \multicolumn{1}{c}{
    {\textcolor{drive. suf.}{$\blacksquare$} drive. suf.~} 
	{\textcolor{other flat}{$\blacksquare$} other flat~} 
	{\textcolor{sidewalk}{$\blacksquare$} sidewalk~} 
	{\textcolor{terrain}{$\blacksquare$} terrain~} 
	{\textcolor{manmade}{$\blacksquare$} manmade~} 
	{\textcolor{vegetation}{$\blacksquare$} vegetation~}		
    }
    \end{tabular}
    \caption{\textbf{Qualitative results of 3D semantic occupancy on the nuScenes-Occupancy validation set.} The input multi-view images are shown on the leftmost side, and the occupancy predictions of the dense baseline C-CONet~\cite{OpenOccupancy}, SparseOcc, SparseOcc++, and the ground-truth are then visualized sequentially. Compared to the 3D dense representation based C-CONet~\cite{yan20222dpass}, our sparse representation achieves better completion and segmentation. The regions highlighted by rectangles show areas with noticeable differences. Best viewed in color and zoomed in.}
    \label{fig:vis_nusc}
    
    \vspace{-2mm}
    
\end{figure*}
\subsection{Qualitative Evaluation}
\label{sec:qualitative_results}
Fig.~\ref{fig:vis_kitti} compares our methods with the dense OccFormer~\cite{occformer} on SemanticKITTI. Our sparse representation minimizes artifacts, effectively eliminating trailing noise from dynamic vehicles (row 1) and reducing hallucinations around parked cars (row 2). Furthermore, SparseOcc++ excels at fine-grained perception, accurately completing small objects such as a bicyclist (row 3) and a crossing pedestrian (row 4) that OccFormer misses or misclassifies.

Fig.~\ref{fig:vis_nusc} extends this comparison to nuScenes-Occupancy against the dense C-CONet~\cite{OpenOccupancy}. Under challenging conditions, including rainy (row 1), night driving (row 2), and crowded environments (row 3), our methods robustly complete both large planar structures (e.g., roads) and dynamic agents. In summary, these results confirm that our geometry-aware sparse representations achieve precise scene completion with superior efficiency compared to dense baselines.

\subsection{Ablation Studies}
\label{sec:ablation_study}
\begin{table}[t]
    \centering
    \caption{Ablation on different designs of Sparse Completion Block on the SemanticKITTI validation set.} 
    \resizebox{0.45\textwidth}{!}{
    \begin{tabular}{c|c|c|c|c}
         \toprule
          Conv Type &  Conv Num & Kernel Size &  IoU & mIoU\\
        \hline
       \noalign{\smallskip}
       - & 0 & - & 35.5 & 12.1 \\
       \hline
       \multirow{3}{*}{Regular}
           & 1 & \multirow{3}{*}{3$\times$3$\times$3} &  35.8 & 12.2 \\
           \cline{2-2}\cline{4-5}
           & 2  & & 36.4 & 12.3\\
           \cline{2-2}\cline{4-5}
           & 3 & & 36.2 & 12.6\\
        \hline
        \multirow{3}{*}{Decomposed}
           &  1  & 3$\times$3$\times$1 & 36.5 & \textbf{13.1}\\ 
           \cline{2-2}\cline{4-5}
           &  2  & 3$\times$1$\times$3 & \textbf{36.6} & 12.8 \\
           \cline{2-2}\cline{4-5}
           &  3  & 1$\times$3$\times1$ & 36.4 & 12.7 \\
         \bottomrule
    \end{tabular}}\
    \label{tab:abl_scb} 
\end{table}
\noindent\textbf{Sparse Completion Block.}
As detailed in Sec.~\ref{sec:sparse_latent_diffuser}, the sparse completion block spatially decomposes a 3D sparse convolution kernel into a combination of three orthogonal convolution kernels. Tab.~\ref{tab:abl_scb} ablates the type and number of sparse kernels. Interestingly, we find that SparseOcc performs satisfactorily even without applying any sparse completion block. We believe this is because the decoder, which fuses the sparse feature pyramid, can also partially complete the scene. Compared with the regular 3$\times$3$\times$3 kernel, the decomposed orthogonal kernels achieve superior scene completion and semantic segmentation results. 
Additionally, stacking more convolutional blocks does not yield performance improvements. Hence, we build our sparse completion block with only a group of decomposed orthogonal kernels. 

\noindent\textbf{Sparse Feature Pyramid.}
\begin{table}[t]
    \centering
    \caption{Ablation on voxel decoder on the SemanticKITTI validation set.} 
    \resizebox{0.45\textwidth}{!}{
    \begin{tabular}{c|c|c|c|c}
         \toprule
          Type  &  IoU & mIoU & Memory & FLOPs\\
        \hline
       FPN3D\cite{fpn} & 34.4 & 9.8 & 13.2G & 307G\\
       MSDeformAttn3D & \textbf{36.7} & \textbf{13.3} & 19.8G & 379G\\
       \rowcolor{gray! 20} Sparse Decoder & 36.5 & 13.1 & 13.3G & 279G\\ 
       \bottomrule
    \end{tabular}}
    \label{tab:abl_vox_dec} 
\end{table}
As shown in Tab.~\ref{tab:abl_vox_dec}, when SparseOcc is equipped with 6 layers of multi-scale deformable attention (MSDeformAttn), it outperforms the proposed sparse decoder. However, as the number of layers in MSDeformAttn decreases, both IoU and mIoU decline, falling behind our simple sparse decoder. Moreover, the training GPU memory and FLOPs of the MSDeformAttn3D are much higher than those of the proposed interpolation and summation method. Compared with FPN3D~\cite{fpn}, our sparse voxel decoder yields a 2.1\% increase in IoU and a 3.3\% increase in mIoU, with only a marginal increase in memory usage.

\begin{table}[t]
  \centering
  \caption{Ablation on segmentation head on the SemanticKITTI validation set.} 
  \resizebox{0.45\textwidth}{!}{
    \begin{tabular}{c|c|c|c|c}
         \toprule
          Type &   IoU & mIoU & Memory & FLOPs\\
        \hline
        Linear Head & \textbf{36.8} & 11.8 & 9.8G & 5.4G\\
        Trans. Head & 36.2 &  \textbf{13.2} & 19.9G & 19.0G\\
        \rowcolor{gray! 20}Sparse Trans. Head  & 36.5 & 13.1 & 13.3G & 13.5G \\ 
           
         \bottomrule
    \end{tabular}}
    \label{tab:abl_trans_dec} 
\end{table}
\noindent\textbf{Sparse Transformer Head.}
We compare several prediction heads, including a simple linear head, the transformer head proposed in~\cite{occformer}, and our sparse transformer head in Tab.~\ref{tab:abl_trans_dec}. As shown, the linear head achieves the highest geometry IoU of 36.8. We hypothesize that this occurs because the linear head is supervised by an explicit geometry loss, i.e., a cross-entropy loss between occupied and non-occupied voxels. Conversely, the transformer decoder formulates occupancy prediction as mask generation for semantic classes only, imposing no explicit supervision on occupied voxels. SparseOcc employs a linear layer for coarse segmentation to filter out non-occupied voxels, followed by a sparse transformer head for mask prediction. Therefore, it strikes an effective balance between these two approaches, achieving satisfactory performance with low training memory requirements.

\begin{table}[t]
  \centering
  \caption{Scaling up the 3D representation resolution on the SemanticKITTI validation set.} 
  \resizebox{0.4\textwidth}{!}{
    \begin{tabular}{c|c|c|c}
         \hline
          Type & 3D resolution &  IoU & mIoU\\
        \hline\hline
        \multirow{2}{*}{SparseOcc-Linear} & 128$\times$128$\times$16 & \textbf{36.8} & 11.8 \\
        
        & 256$\times$256$\times$256 & 36.4 & \textbf{12.3} \\ 
           
         \hline
    \end{tabular}}
    \label{tab:abl_3d_res} 
\end{table}
\noindent\textbf{3D Resolution.}
SparseOcc utilizes LSS~\cite{lss} to lift 2D image features into a 3D volume, from which the 3D sparse representation is extracted. The resolution of the LSS output volume may influence performance. To investigate this efficiently, we use SparseOcc with a linear segmentation head. As shown in Tab.~\ref{tab:abl_3d_res}, scaling up the LSS output resolution by 2$\times$ boosts mIoU from 11.8 to 12.3 but decreases geometry IoU by 0.4. We attribute this IoU drop to the insufficient number of sparse completion blocks to cover the entire scene at a higher resolution. This issue can be resolved by stacking an additional sparse completion block at the final layer of the 3D sparse encoder.

\begin{table}[t]
  \centering
  \caption{Ablation on the components of geometry-aware sparse representation on the SemanticKITTI validation set.} 
  \resizebox{0.45\textwidth}{!}{
    \begin{tabular}{ccc|c|c}
        \toprule
        \makecell{Sparse Anchor \\ Voxel} & \makecell{Compact Scene \\ Completion } & \makecell{Constrained \\ Sem. Pre.} & mIoU & FPS \\
        \midrule
        &          &            &    11.32   & 5.3  \\
        \checkmark &            &            &     8.9    & 28.4  \\
        \checkmark & \checkmark &            &     11.54  & 18.3   \\
        \rowcolor{gray! 20}\checkmark & \checkmark & \checkmark & 
             13.23  & 24.7   \\
        \bottomrule
        \end{tabular}}
    \label{tab:component} 
\end{table}
\noindent\textbf{Geometry-Aware Sparse Representation.}
We explore the performance and speed gains contributed by each module of our geometry-aware sparse representation in Tab.~\ref{tab:component}. When all modules are disabled, the model degrades to a dense version, achieving competitive mIoU but at a reduced speed. When utilizing only sparse anchor voxel generation without scene completion, performance drops significantly due to compromised scene integrity. However, adding compact scene completion increases IoU, albeit with a speed reduction due to the processing of newly introduced voxels. By constraining semantic prediction to the completed scene, we achieve the optimal mIoU-FPS trade-off.

\begin{table}[t]
  \centering
  \caption{Different types of sparse anchor voxel generation on the SemanticKITTI validation set.} 
  \resizebox{0.35\textwidth}{!}{
    \begin{tabular}{c|c|c|c}
         \toprule
          No. & Type  &  IoU & mIoU  \\
        \midrule
       (1) & Uniform Sampling & 36.82 & 11.83  \\
       \rowcolor{gray! 20} (2) &  Coarse-to-Fine & 37.08 & 13.23 \\ 
       \bottomrule
    \end{tabular}}
    \label{tab:abl_sav} 
\end{table}
\noindent\textbf{Sparse Anchor Voxel Generation.} 
We compare the coarse-to-fine strategy with uniform sampling in Tab.~\ref{tab:abl_sav}. We find that the coarse-to-fine approach yields better performance, as it consistently focuses on the most promising occupancy priors.

\begin{table}[t]
  \centering
  \caption{Ablation on directional distance prediction.}\label{tab:abl_dist_pred}
  \resizebox{0.47\textwidth}{!}{
    \begin{tabular}{ccc|c|c}
          \toprule
          \multirow{2}{*}{\shortstack{Orthogonal \\ Decomposition}} & \multirow{2}{*}{\shortstack{Discretized \\ Learning}} & \multirow{2}{*}{\shortstack{Geometry-Guided \\Propagation}} & \multirow{2}{*}{mIoU} & \multirow{2}{*}{FPS} \\
           &  &  & \\
          \midrule
            &          &     & 10.31   & 15.8  \\
            \checkmark &     &   & 12.36  & 15.7   \\
            \checkmark & \checkmark &   & 12.72  & 15.7   \\
            \rowcolor{gray! 20}\checkmark & \checkmark & \checkmark  & 13.23  & 24.7   \\
            \bottomrule
        \end{tabular}
        }
\end{table}
\noindent\textbf{Scene Completion Field Learning Strategies.}
We validate the design choices for SCF learning strategies in Tab.~\ref{tab:abl_dist_pred}. Given that outdoor scenes exhibit strong geometric anisotropy, the baseline model fails to accurately complete the scene with a single scalar SCF, achieving an mIoU of only 10.31. Introducing orthogonal decomposition significantly improves performance by 2.05 mIoU. This suggests that decomposing the SCF into independent orthogonal components simplifies the learning task. Furthermore, incorporating discretized learning yields an additional 0.87 mIoU gain. Moreover, replacing the sparse latent propagator, which employs indiscriminate feature propagation, with our geometry-guided propagation further increases mIoU to 13.23 and improves inference speed from 15.7 to 24.7 FPS, demonstrating the efficiency of our proposed SCF learning strategies.

\begin{table}[t]
  \centering
  \caption{Ablation on compact scene completion.} 
  \resizebox{0.34\textwidth}{!}
  {
    \begin{tabular}{c|c|c|c}
         \midrule
        \makecell{\# Scene \\ Completion}  & $({s}_{\text{max}}^{\text{X-Y}},{s}_{\text{max}}^{\text{Z}})$  & IoU & mIoU  \\
        \midrule
       0 & - & 33.83 & 8.9 \\
       \midrule
       1 & $(5, 3)$ & 36.66 & 11.50 \\
       \rowcolor{gray! 20} 1 & $(10, 3)$ & 37.08  & 13.23  \\
       \midrule
       \rowcolor{gray! 20}2 & $(5, 3)$ & 37.19 & 13.17\\
       2 & $(10, 3)$ &  37.46 & 12.08 \\
       \midrule
    \end{tabular}}
    \label{tab:abl_gr} 
\end{table}
\noindent\textbf{Compact Scene Completion.} 
The scene completion operation can be applied iteratively, where occupied voxels generated in one step serve as sparse anchors for the next, thereby progressively increasing the density of the predicted space. 
Additionally, the maximum SCF vector $({s}_{\text{max}}^{\text{X-Y}},{s}_{\text{max}}^{\text{Z}})$ directly governs the range of voxel propagation.
To investigate the impact of the iteration count and the maximum SCF distance, we conduct an ablation study across different settings, as shown in Tab.~\ref{tab:abl_gr}. 
When the iteration count is zero (i.e., no completion), the network degrades to a naive dense-to-sparse pruning baseline, destroying structural integrity and yielding suboptimal accuracy. 
Increasing the number of iterations initially improves IoU by recovering more complete geometry. 
However, we observe that applying the operation twice (2 iterations) with a large range of $(10, 3)$ reduces the mIoU to 12.08, likely due to error accumulation or geometric ambiguity arising from over-propagation. 
Consequently, to balance accuracy and efficiency, we adopt a single-iteration setting with the maximum SCF distance set to $(10, 3)$.

\begin{table}[t]
  \centering
  \caption{Ablation on input size and 2D backbone.} 
  \resizebox{0.47\textwidth}{!}{
    \begin{tabular}{c|c|c|c|c}
         \midrule
          Method &  2D Backbone &  Input Size & IoU & mIoU\\
          \midrule
          \multicolumn{5}{c}{\textit{nuScenes-Occupancy}~\cite{OpenOccupancy}} \\
        \midrule
       
       C-CONet~\cite{OpenOccupancy} & R-50 & $704\times 256$ &16.6 &8.6 \\ 
       C-CONet~\cite{OpenOccupancy} & R-50 & $1600\times 900$ & 19.3 & 10.3\\
       \rowcolor{gray! 20}C-CONet~\cite{OpenOccupancy} & R-101 & $1600\times 900$ & 20.2 & 11.4\\
        \midrule
        SparseOcc &  R-50  & $704\times 256$ &  {21.8} & 14.1\\ 
          
        SparseOcc &  R-50  & $1600\times 900$ & 20.4 & {14.6} \\      
        SparseOcc++ &  R-50  & $704\times 256$ &  24.1 & 14.1\\ 
        \rowcolor{gray! 20}SparseOcc++ &  R-50  & $1600\times 900$ & 24.5 & 14.01 \\   
         \midrule
          \multicolumn{5}{c}{\textit{SemanticKITTI}~\cite{semantickitti}} \\
        \midrule
        SparseOcc & R-50 & $384 \times 1280$ & 36.48 & 13.12 \\
        SparseOcc++   &  R-50  & $384 \times 1280$ & 37.08 & 13.23   \\ 
          
        \rowcolor{gray! 20}SparseOcc++ &  Swin-B  & $384 \times 1280$ & 37.23 & 13.97 \\   
         \hline
    \end{tabular}}
    \label{tab:abl_img_bb} 
\end{table}
\noindent\textbf{Image Backbone and Input Size.} 
We study the influence of different image backbones and input sizes in Tab.~\ref{tab:abl_img_bb}. SparseOcc outperforms C-CONet on the nuScenes-Occupancy dataset even when trained with a smaller input size ($704\times256$) and a lighter backbone (ResNet-50), demonstrating the effectiveness of the architecture. Additionally, using a larger input size ($1600\times900$) improves mIoU for both C-CONet and SparseOcc due to the increased density of semantic features.
However, the IoU of SparseOcc decreases at higher image resolutions. We attribute this to the erroneous hallucination of spatially empty voxels caused by over-dense 3D sparse features. 
Moreover, our SparseOcc++ achieves the best IoU by a significant margin (24.1 {vs.} 20.2 on nuScenes-Occupancy), even with a smaller image backbone and input size, demonstrating its superior design. Furthermore, a stronger image encoder, such as Swin Transformer~\cite{swin_transformer}, could further improve network performance.

\begin{figure}[t]
    \centering
    \includegraphics[scale=.7]{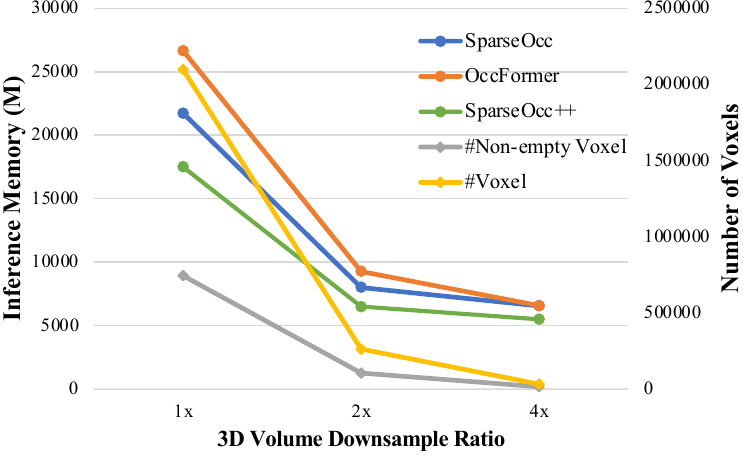}
    \caption{\textbf{Efficiency analysis when scaling up the 3D representation on SemanticKITTI~\cite{semantickitti} validation set.} The left axis represents the inference GPU memory. The right axis denotes the number of voxels. The 3D downsampling ratio is considered between the ground-truth and the LSS output. The number of non-empty voxels is measured using max-donwsampled ground-truth.}
    \label{fig:infer_memory}
    \vspace{-2mm}
\end{figure}
\noindent\textbf{Efficiency Analysis.}
\label{sec:efficiency_analysis}
The complexity of our sparse representations is expected to be approximately linear with respect to the number of non-empty voxels, as operations are restricted to feature-occupied regions. We compare the inference GPU memory and the number of non-empty voxels in Fig.~\ref{fig:infer_memory}. As observed, when scaling up the 3D resolution from $2\times$ to $1\times$, the dense representation-based method OccFormer~\cite{occformer} suffers from a steep increase in GPU memory usage. In contrast, SparseOcc and SparseOcc++ exhibit a linear increase relative to the number of non-empty voxels, confirming the superior efficiency of the 3D sparse representation.
\section{Limitations} \label{sec:limitations} Despite its efficiency, SparseOcc++ has two primary limitations: \textit{(1) Geometric simplification:} The orthogonal directional distance model favors locally convex or structurally simple geometries. For non-convex objects or intricate structures (e.g., irregular `vegetation' in Fig.~\ref{fig:vis_kitti}), our regression-based reconstruction may yield over-simplified results. \textit{(2) Long-range ambiguity:} As a vision-based approach, accurately regressing distances for far-range objects remains ill-posed without direct depth measurements (e.g., long-range `drivable surface' in Fig.~\ref{fig:vis_nusc}). Consequently, geometric precision degrades at a distance compared to multi-modal solutions, {i.e.}, 24.1 (SparseOcc++) {vs.} 30.3 (OccGen~\cite{wang2024occgen}). Future research will focus on three directions: (1) incorporating temporal cues to resolve long-range depth ambiguity and ensure cross-frame consistency; (2) exploring adaptive anchor sampling to better capture complex, non-convex geometries; and (3) extending the framework to multi-modal fusion to leverage the geometric robustness of LiDAR while maintaining the efficiency of sparse representations.

\section{Conclusion}
In this paper, we present SparseOcc++, a comprehensive extension of our preliminary work, SparseOcc, designed to address efficiency bottlenecks in vision-based 3D occupancy prediction. 
We identify that existing sparse methods suffer from an entanglement of scene completion and semantic classification, which leads to computational redundancy in empty regions and boundary ambiguity. 
To address this limitation, we propose a novel geometry-aware sparse representation that explicitly disentangles these two processes. 
By reformulating scene completion as a scene completion field (SCF) learning task defined on sparse anchor voxels, we successfully reduce unnecessary computation while enhancing structural fidelity.
To implement the SCF effectively, we introduce an orthogonal decomposition strategy and a discretized learning scheme for robust anisotropic modeling, followed by a geometry-guided voxel propagation mechanism that efficiently reconstructs the explicit volumetric structure.
Extensive experiments on the nuScenes and SemanticKITTI datasets demonstrate that SparseOcc++ achieves state-of-the-art performance, showing superior IoU and faster inference speeds compared to both its predecessor and dense counterparts. 
We believe that this fully sparse geometry-aware representation offers a scalable and efficient direction for future 3D perception in autonomous driving.

{
\bibliographystyle{IEEEtran}
\bibliography{main}

@String(PAMI = {IEEE Trans. Pattern Anal. Mach. Intell.})

@String(CVPR= {IEEE Conf. Comput. Vis. Pattern Recog.})

@String(ICCV= {Int. Conf. Comput. Vis.})

@String(ECCV= {Eur. Conf. Comput. Vis.})

@String(NIPS= {Adv. Neural Inform. Process. Syst.})

@String(TOG= {ACM Trans. Graph.})

@String(ICLR = {Int. Conf. Learn. Represent.})

@String(AAAI = {AAAI})

@String(PAMI  = {IEEE TPAMI})

@String(CVPR  = {CVPR})

@String(ICCV  = {ICCV})

@String(ECCV  = {ECCV})

@String(NIPS  = {NeurIPS})

@String(TOG   = {ACM TOG})

@String(TCSVT = {IEEE TCSVT})

@String(ICRA  = {ICRA})

@String(ICLR  = {ICLR})

@inproceedings{unet,
  title={U-net: Convolutional networks for biomedical image segmentation},
  author={Ronneberger, Olaf and Fischer, Philipp and Brox, Thomas},
  booktitle={MICCAI},
  year={2015},
}

@inproceedings{jiang2024symphonize,
  title={Symphonize 3d semantic scene completion with contextual instance queries},
  author={Jiang, Haoyi and Cheng, Tianheng and Gao, Naiyu and Zhang, Haoyang and Lin, Tianwei and Liu, Wenyu and Wang, Xinggang},
  booktitle=CVPR,
  year={2024}
}

@inproceedings{yao2023ndc,
  title={Ndc-scene: Boost monocular 3d semantic scene completion in normalized device coordinates space},
  author={Yao, Jiawei and Li, Chuming and Sun, Keqiang and Cai, Yingjie and Li, Hao and Ouyang, Wanli and Li, Hongsheng},
  booktitle=ICCV,
  year={2023},
}

@article{zhang2023occnerf,
  title={Occnerf: Self-supervised multi-camera occupancy prediction with neural radiance fields},
  author={Zhang, Chubin and Yan, Juncheng and Wei, Yi and Li, Jiaxin and Liu, Li and Tang, Yansong and Duan, Yueqi and Lu, Jiwen},
  journal={arXiv:2312.09243},
  year={2023}
}

@inproceedings{pan2024renderocc,
  title={Renderocc: Vision-centric 3d occupancy prediction with 2d rendering supervision},
  author={Pan, Mingjie and Liu, Jiaming and Zhang, Renrui and Huang, Peixiang and Li, Xiaoqi and Xie, Hongwei and Wang, Bing and Liu, Li and Zhang, Shanghang},
  booktitle=ICRA,
  year={2024},
}

@inproceedings{zhang2024safdnet,
  title={SAFDNet: A Simple and Effective Network for Fully Sparse 3D Object Detection},
  author={Zhang, Gang and Chen, Junnan and Gao, Guohuan and Li, Jianmin and Liu, Si and Hu, Xiaolin},
  booktitle=CVPR,
  year={2024}
}

@inproceedings{lidarrcnn,
  title={Lidar r-cnn: An efficient and universal 3d object detector},
  author={Li, Zhichao and Wang, Feng and Wang, Naiyan},
  booktitle=CVPR,
  year={2021}
}

@inproceedings{park2019deepsdf,
  title={Deepsdf: Learning continuous signed distance functions for shape representation},
  author={Park, Jeong Joon and Florence, Peter and Straub, Julian and Newcombe, Richard and Lovegrove, Steven},
  booktitle=CVPR,
  year={2019}
}

@article{lu2023octreeocc,
  title={OctreeOcc: Efficient and multi-granularity occupancy prediction using octree queries},
  author={Lu, Yuhang and Zhu, Xinge and Wang, Tai and Ma, Yuexin},
  journal={arXiv:2312.03774},
  year={2023}
}

@article{huang2024gaussianformer,
  title={GaussianFormer: Scene as Gaussians for Vision-Based 3D Semantic Occupancy Prediction},
  author={Huang, Yuanhui and Zheng, Wenzhao and Zhang, Yunpeng and Zhou, Jie and Lu, Jiwen},
  journal={arXiv:2405.17429},
  year={2024}
}

@article{pan2023uniocc_xiaomi,
  title={Uniocc: Unifying vision-centric 3d occupancy prediction with geometric and semantic rendering},
  author={Pan, Mingjie and Liu, Li and Liu, Jiaming and Huang, Peixiang and Wang, Longlong and Zhang, Shanghang and Xu, Shaoqing and Lai, Zhiyi and Yang, Kuiyuan},
  journal={arXiv:2306.09117},
  year={2023}
}

@article{yu2023flashocc,
  title={Flashocc: Fast and memory-efficient occupancy prediction via channel-to-height plugin},
  author={Yu, Zichen and Shu, Changyong and Deng, Jiajun and Lu, Kangjie and Liu, Zongdai and Yu, Jiangyong and Yang, Dawei and Li, Hui and Chen, Yan},
  journal={arXiv:2311.12058},
  year={2023}
}

@inproceedings{chen2025alocc,
  title={Alocc: Adaptive lifting-based 3d semantic occupancy and cost volume-based flow predictions},
  author={Chen, Dubing and Fang, Jin and Han, Wencheng and Cheng, Xinjing and Yin, Junbo and Xu, Chengzhong and Khan, Fahad Shahbaz and Shen, Jianbing},
  booktitle=ICCV,
  year={2025}
}

@inproceedings{liu2023fully,
  title={Fully sparse 3d panoptic occupancy prediction},
  author={Liu, Haisong and Wang, Haiguang and Chen, Yang and Yang, Zetong and Zeng, Jia and Chen, Li and Wang, Limin},
  booktitle=ECCV,
  year={2024}
}

@inproceedings{wang2024panoocc,
  title={Panoocc: Unified occupancy representation for camera-based 3d panoptic segmentation},
  author={Wang, Yuqi and Chen, Yuntao and Liao, Xingyu and Fan, Lue and Zhang, Zhaoxiang},
  booktitle=CVPR,
  year={2024}
}

@inproceedings{tang2024sparseocc,
  title={Sparseocc: Rethinking sparse latent representation for vision-based semantic occupancy prediction},
  author={Tang, Pin and Wang, Zhongdao and Wang, Guoqing and Zheng, Jilai and Ren, Xiangxuan and Feng, Bailan and Ma, Chao},
  booktitle=CVPR,
  year={2024}
}

@inproceedings{zheng2024veon,
  title={VEON: Vocabulary-Enhanced Occupancy Prediction},
  author={Zheng, Jilai and Tang, Pin and Wang, Zhongdao and Wang, Guoqing and Ren, Xiangxuan and Feng, Bailan and Ma, Chao},
  booktitle=ECCV,
  year={2024}
}

@inproceedings{wang2024occgen,
  title={Occgen: Generative multi-modal 3d occupancy prediction for autonomous driving},
  author={Wang, Guoqing and Wang, Zhongdao and Tang, Pin and Zheng, Jilai and Ren, Xiangxuan and Feng, Bailan and Ma, Chao},
  booktitle=ECCV,
  year={2024}
}

@article{fbocc,
  title={Fb-occ: 3d occupancy prediction based on forward-backward view transformation},
  author={Li, Zhiqi and Yu, Zhiding and Austin, David and Fang, Mingsheng and Lan, Shiyi and Kautz, Jan and Alvarez, Jose M},
  journal={arXiv:2307.01492},
  year={2023}
}

@inproceedings{vobecky2024pop3d,
  title={Pop-3d: Open-vocabulary 3d occupancy prediction from images},
  author={Vobecky, Antonin and Sim{\'e}oni, Oriane and Hurych, David and Gidaris, Spyridon and Bursuc, Andrei and P{\'e}rez, Patrick and Sivic, Josef},
  booktitle=NIPS,
  year={2023}
}

@inproceedings{semantickitti,
  title={Semantickitti: A dataset for semantic scene understanding of lidar sequences},
  author={Behley, Jens and Garbade, Martin and Milioto, Andres and Quenzel, Jan and Behnke, Sven and Stachniss, Cyrill and Gall, Jurgen},
  booktitle=ICCV,
  year={2019}
}

@INPROCEEDINGS{waymo,
  title={Scalability in Perception for Autonomous Driving: Waymo Open Dataset},
  author={Pei Sun and Henrik Kretzschmar and Xerxes Dotiwalla and Aurelien Chouard and Vijaysai Patnaik and Paul Tsui and James Guo and Yin Zhou and Yuning Chai and Benjamin Caine and Vijay K. Vasudevan and Wei Han and Jiquan Ngiam and Hang Zhao and Aleksei Timofeev and Scott Ettinger and Maxim Krivokon and Amy Gao and Aditya Joshi and Yu Zhang and Jonathon Shlens and Zhifeng Chen and Dragomir Anguelov},
  booktitle = {CVPR},
  year={2020}
}

@inproceedings{js3cnet,
  title={Sparse single sweep lidar point cloud segmentation via learning contextual shape priors from scene completion},
  author={Yan, Xu and Gao, Jiantao and Li, Jie and Zhang, Ruimao and Li, Zhen and Huang, Rui and Cui, Shuguang},
  booktitle={AAAI},
  year={2021}
}

@article{LDIF,
  title={Semantic scene completion using local deep implicit functions on lidar data},
  author={Rist, Christoph B and Emmerichs, David and Enzweiler, Markus and Gavrila, Dariu M},
  journal=PAMI,
  volume={44},
  number={10},
  pages={7205--7218},
  year={2021},
  publisher={IEEE}
}

@inproceedings{monoscene,
  title={Monoscene: Monocular 3d semantic scene completion},
  author={Cao, Anh-Quan and de Charette, Raoul},
  booktitle={CVPR},
  year={2022}
}

@inproceedings{zhou2018voxelnet,
  title={Voxelnet: End-to-end learning for point cloud based 3d object detection},
  author={Zhou, Yin and Tuzel, Oncel},
  booktitle={CVPR},
  year={2018}
}

@inproceedings{fpn,
  title={Feature pyramid networks for object detection},
  author={Lin, Tsung-Yi and Doll{\'a}r, Piotr and Girshick, Ross and He, Kaiming and Hariharan, Bharath and Belongie, Serge},
  booktitle={CVPR},
  year={2017}
}

@inproceedings{lss,
  title={Lift, splat, shoot: Encoding images from arbitrary camera rigs by implicitly unprojecting to 3d},
  author={Philion, Jonah and Fidler, Sanja},
  booktitle={ECCV},
  year={2020},
}

@inproceedings{lovasz,
  title={The lov{\'a}sz-softmax loss: A tractable surrogate for the optimization of the intersection-over-union measure in neural networks},
  author={Berman, Maxim and Triki, Amal Rannen and Blaschko, Matthew B},
  booktitle={CVPR},
  year={2018}
}

@article{bevfusion,
  title={BEVFusion: Multi-Task Multi-Sensor Fusion with Unified Bird's-Eye View Representation},
  author={Liu, Zhijian and Tang, Haotian and Amini, Alexander and Yang, Xinyu and Mao, Huizi and Rus, Daniela and Han, Song},
  journal={arXiv:2205.13542},
  year={2022}
}

@inproceedings{pointpillar,
  title={Pointpillars: Fast encoders for object detection from point clouds},
  author={Lang, Alex H and Vora, Sourabh and Caesar, Holger and Zhou, Lubing and Yang, Jiong and Beijbom, Oscar},
  booktitle={CVPR},
  year={2019}
}

@InProceedings{OpenOccupancy,
    author    = {Wang, Xiaofeng and Zhu, Zheng and Xu, Wenbo and Zhang, Yunpeng and Wei, Yi and Chi, Xu and Ye, Yun and Du, Dalong and Lu, Jiwen and Wang, Xingang},
    title     = {OpenOccupancy: A Large Scale Benchmark for Surrounding Semantic Occupancy Perception},
    booktitle = {ICCV},
    year      = {2023}
}

@inproceedings{SCPNet,
  title={SCPNet: Semantic Scene Completion on Point Cloud},
  author={Xia, Zhaoyang and Liu, Youquan and Li, Xin and Zhu, Xinge and Ma, Yuexin and Li, Yikang and Hou, Yuenan and Qiao, Yu},
  booktitle={CVPR},
  year={2023}
}

@inproceedings{Cylinder3D,
  title={Cylindrical and asymmetrical 3d convolution networks for lidar segmentation},
  author={Zhu, Xinge and Zhou, Hui and Wang, Tai and Hong, Fangzhou and Ma, Yuexin and Li, Wei and Li, Hongsheng and Lin, Dahua},
  booktitle={CVPR},
  year={2021}
}

@misc{spconv,
    title={Spconv: Spatially Sparse Convolution Library},
    author={Spconv Contributors},
    howpublished = {\url{https://github.com/traveller59/spconv}},
    year={2022}
}

@inproceedings{mask2former,
  title={Masked-attention mask transformer for universal image segmentation},
  author={Cheng, Bowen and Misra, Ishan and Schwing, Alexander G and Kirillov, Alexander and Girdhar, Rohit},
  booktitle={CVPR},
  year={2022}
}

@inproceedings{occformer,
  title={Occformer: Dual-path transformer for vision-based 3d semantic occupancy prediction},
  author={Zhang, Yunpeng and Zhu, Zheng and Du, Dalong},
  booktitle=ICCV,
  year={2023}
}

@inproceedings{deformabledetr,
  title={Deformable DETR: Deformable Transformers for End-to-End Object Detection},
  author={Zhu, Xizhou and Su, Weijie and Lu, Lewei and Li, Bin and Wang, Xiaogang and Dai, Jifeng},
  booktitle={ICLR},
  year={2020}
}

@article{bevdet,
  title={Bevdet: High-performance multi-camera 3d object detection in bird-eye-view},
  author={Huang, Junjie and Huang, Guan and Zhu, Zheng and Du, Dalong},
  journal={arXiv:2112.11790},
  year={2021}
}

@inproceedings{BEVFormer,
  title={Bevformer: Learning bird’s-eye-view representation from multi-camera images via spatiotemporal transformers},
  author={Li, Zhiqi and Wang, Wenhai and Li, Hongyang and Xie, Enze and Sima, Chonghao and Lu, Tong and Qiao, Yu and Dai, Jifeng},
  booktitle=ECCV,
  year={2022},
}

@InProceedings{deformable_DETR,
  title     = {Deformable {DETR}: Deformable Transformers for End-to-End Object Detection},
  author    = {Xizhou Zhu and Weijie Su and Lewei Lu and Bin Li and Xiaogang Wang and Jifeng Dai},
  booktitle = {ICLR},
  year      = {2021},
}

@inproceedings{sscnet,
  title={Semantic scene completion from a single depth image},
  author={Song, Shuran and Yu, Fisher and Zeng, Andy and Chang, Angel X and Savva, Manolis and Funkhouser, Thomas},
  booktitle={CVPR},
  year={2017}
}

@inproceedings{swin_transformer,
  title={Swin transformer: Hierarchical vision transformer using shifted windows},
  author={Liu, Ze and Lin, Yutong and Cao, Yue and Hu, Han and Wei, Yixuan and Zhang, Zheng and Lin, Stephen and Guo, Baining},
  booktitle={ICCV},
  year={2021}
}

@article{bevsegformer,
  title={BEVSegFormer: Bird's Eye View Semantic Segmentation From Arbitrary Camera Rigs},
  author={Peng, Lang and Chen, Zhirong and Fu, Zhangjie and Liang, Pengpeng and Cheng, Erkang},
  journal={arXiv:2203.04050},
  year={2022}
}

@InProceedings{nuscenes,
  title={Nuscenes: A multimodal dataset for autonomous driving},
  author={Caesar, Holger and Bankiti, Varun and Lang, Alex H and Vora, Sourabh and Liong, Venice Erin and Xu, Qiang and Krishnan, Anush and Pan, Yu and Baldan, Giancarlo and Beijbom, Oscar},
  booktitle={CVPR},
  year={2020}
}

@InProceedings{faster_rcnn,
  author    = {Ren, Shaoqing and He, Kaiming  and Girshick, Ross and Sun, Jian },
  title     = {Faster {R-CNN}: Towards Real-Time Object Detection with Region Proposal Networks},
  booktitle = {NeurIPS},
  year      = {2015}
}

@InProceedings{voxelnet,
  title={Voxnet: A 3d convolutional neural network for real-time object recognition},
  author={Maturana, Daniel and Scherer, Sebastian},
  booktitle={IROS},
  year={2015},
}

@article{second,
  title={Second: Sparsely embedded convolutional detection},
  author={Yan, Yan and Mao, Yuxing and Li, Bo},
  journal={Sensors},
  year={2018},
  publisher={Multidisciplinary Digital Publishing Institute}
}

@InProceedings{resnet,
  title={Deep residual learning for image recognition},
  author={He, Kaiming and Zhang, Xiangyu and Ren, Shaoqing and Sun, Jian},
  booktitle={CVPR},
  year={2016}
}

@inproceedings{efficientnet,
  title={Efficientnet: Rethinking model scaling for convolutional neural networks},
  author={Tan, Mingxing and Le, Quoc},
  booktitle={ICML},
  year={2019},
}

@inproceedings{tpvformer,
  title={Tri-perspective view for vision-based 3d semantic occupancy prediction},
  author={Huang, Yuanhui and Zheng, Wenzhao and Zhang, Yunpeng and Zhou, Jie and Lu, Jiwen},
  booktitle=CVPR,
  year={2023}
}

@inproceedings{roldao2020lmscnet,
  title={Lmscnet: Lightweight multiscale 3d semantic completion},
  author={Roldao, Luis and de Charette, Raoul and Verroust-Blondet, Anne},
  booktitle={3DV},
  year={2020},
}

@inproceedings{3d-sketch,
  title={3d sketch-aware semantic scene completion via semi-supervised structure prior},
  author={Chen, Xiaokang and Lin, Kwan-Yee and Qian, Chen and Zeng, Gang and Li, Hongsheng},
  booktitle={CVPR},
  year={2020}
}

@inproceedings{li2020anisotropic,
  title={Anisotropic convolutional networks for 3d semantic scene completion},
  author={Li, Jie and Han, Kai and Wang, Peng and Liu, Yu and Yuan, Xia},
  booktitle={CVPR},
  year={2020}
}

@inproceedings{hu2021fiery,
  title={FIERY: future instance prediction in bird's-eye view from surround monocular cameras},
  author={Hu, Anthony and Murez, Zak and Mohan, Nikhil and Dudas, Sof{\'\i}a and Hawke, Jeffrey and Badrinarayanan, Vijay and Cipolla, Roberto and Kendall, Alex},
  booktitle={ICCV},
  year={2021}
}

@inproceedings{pon,
  title={Predicting semantic map representations from images using pyramid occupancy networks},
  author={Roddick, Thomas and Cipolla, Roberto},
  booktitle={CVPR},
  year={2020}
}

@inproceedings{CVT,
  title={Cross-view transformers for real-time map-view semantic segmentation},
  author={Zhou, Brady and Kr{\"a}henb{\"u}hl, Philipp},
  booktitle={CVPR},
  year={2022}
}

@article{bevdepth,
  title={Bevdepth: Acquisition of reliable depth for multi-view 3d object detection},
  author={Li, Yinhao and Ge, Zheng and Yu, Guanyi and Yang, Jinrong and Wang, Zengran and Shi, Yukang and Sun, Jianjian and Li, Zeming},
  journal={arXiv:2206.10092},
  year={2022}
}

@inproceedings{surroundocc,
  title={Surroundocc: Multi-camera 3d occupancy prediction for autonomous driving},
  author={Wei, Yi and Zhao, Linqing and Zheng, Wenzhao and Zhu, Zheng and Zhou, Jie and Lu, Jiwen},
  booktitle={CVPR},
  year={2023}
}

@inproceedings{metabev,
  title={Metabev: Solving sensor failures for 3d detection and map segmentation},
  author={Ge, Chongjian and Chen, Junsong and Xie, Enze and Wang, Zhongdao and Hong, Lanqing and Lu, Huchuan and Li, Zhenguo and Luo, Ping},
  booktitle={ICCV},
  year={2023}
}

@inproceedings{uniad,
  title={Planning-oriented autonomous driving},
  author={Hu, Yihan and Yang, Jiazhi and Chen, Li and Li, Keyu and Sima, Chonghao and Zhu, Xizhou and Chai, Siqi and Du, Senyao and Lin, Tianwei and Wang, Wenhai and others},
  booktitle={CVPR},
  year={2023}
}

@inproceedings{rclane,
  title={RCLane: Relay chain prediction for lane detection},
  author={Xu, Shenghua and Cai, Xinyue and Zhao, Bin and Zhang, Li and Xu, Hang and Fu, Yanwei and Xue, Xiangyang},
  booktitle={ECCV},
  year={2022},
}

@inproceedings{tang2023prototransfer,
  title={ProtoTransfer: Cross-Modal Prototype Transfer for Point Cloud Segmentation},
  author={Tang, Pin and Xu, Hai-Ming and Ma, Chao},
  booktitle={ICCV},
  year={2023}
}

@inproceedings{yan20222dpass,
  title={2dpass: 2d priors assisted semantic segmentation on lidar point clouds},
  author={Yan, Xu and Gao, Jiantao and Zheng, Chaoda and Zheng, Chao and Zhang, Ruimao and Cui, Shuguang and Li, Zhen},
  booktitle={ECCV},
  year={2022},
}

@inproceedings{voxformer,
  title={Voxformer: Sparse voxel transformer for camera-based 3d semantic scene completion},
  author={Li, Yiming and Yu, Zhiding and Choy, Christopher and Xiao, Chaowei and Alvarez, Jose M and Fidler, Sanja and Feng, Chen and Anandkumar, Anima},
  booktitle={CVPR},
  year={2023}
}

@inproceedings{tong2023scene,
  title={Scene as occupancy},
  author={Tong, Wenwen and Sima, Chonghao and Wang, Tai and Chen, Li and Wu, Silei and Deng, Hanming and Gu, Yi and Lu, Lewei and Luo, Ping and Lin, Dahua and others},
  booktitle={ICCV},
  year={2023}
}

@article{tian2023occ3d,
  title={Occ3d: A large-scale 3d occupancy prediction benchmark for autonomous driving},
  author={Tian, Xiaoyu and Jiang, Tao and Yun, Longfei and Wang, Yue and Wang, Yilun and Zhao, Hang},
  journal={arXiv:2304.14365},
  year={2023}
}

@article{zuo2023pointocc,
  title={PointOcc: Cylindrical Tri-Perspective View for Point-based 3D Semantic Occupancy Prediction},
  author={Zuo, Sicheng and Zheng, Wenzhao and Huang, Yuanhui and Zhou, Jie and Lu, Jiwen},
  journal={arXiv:2308.16896},
  year={2023}
}

@inproceedings{jia2023driveadapter,
  title={Driveadapter: Breaking the coupling barrier of perception and planning in end-to-end autonomous driving},
  author={Jia, Xiaosong and Gao, Yulu and Chen, Li and Yan, Junchi and Liu, Patrick Langechuan and Li, Hongyang},
  booktitle={CVPR},
  year={2023}
}

@inproceedings{zhang2024alleviating,
  title={Alleviating Foreground Sparsity for Semi-Supervised Monocular 3D Object Detection},
  author={Zhang, Weijia and Liu, Dongnan and Ma, Chao and Cai, Weidong},
  booktitle={WACV},
  year={2024}
}

@inproceedings{zhou2023unidistill,
  title={UniDistill: A Universal Cross-Modality Knowledge Distillation Framework for 3D Object Detection in Bird's-Eye View},
  author={Zhou, Shengchao and Liu, Weizhou and Hu, Chen and Zhou, Shuchang and Ma, Chao},
  booktitle={CVPR},
  year={2023}
}

@inproceedings{wang2025opus,
  title={Opus: occupancy prediction using a sparse set},
  author={Wang, Jiabao and Liu, Zhaojiang and Meng, Qiang and Yan, Liujiang and Wang, Ke and Yang, Jie and Liu, Wei and Hou, Qibin and Cheng, Ming-Ming},
  booktitle={NeurIPS},
  year={2024}
}

@article{gaussianformer2,
  title={Probabilistic Gaussian Superposition for Efficient 3D Occupancy Prediction},
  author={Huang, Yuanhui and Thammatadatrakoon, Amonnut and Zheng, Wenzhao and Zhang, Yunpeng and Du, Dalong and Lu, Jiwen},
  journal={arXiv:2412.04384},
  year={2024}
}

@inproceedings{hou2024fastocc,
  title={Fastocc: Accelerating 3d occupancy prediction by fusing the 2d bird’s-eye view and perspective view},
  author={Hou, Jiawei and Li, Xiaoyan and Guan, Wenhao and Zhang, Gang and Feng, Di and Du, Yuheng and Xue, Xiangyang and Pu, Jian},
  booktitle=ICRA,
  year={2024},
}

@inproceedings{fsd,
  title={Fully sparse 3d object detection},
  author={Fan, Lue and Wang, Feng and Wang, Naiyan and Zhang, Zhao-Xiang},
  booktitle={NeurIPS},
  year={2022}
}

@article{fsdv1,
  title={Super sparse 3d object detection},
  author={Fan, Lue and Yang, Yuxue and Wang, Feng and Wang, Naiyan and Zhang, Zhaoxiang},
  journal=PAMI,
  volume={45},
  number={10},
  pages={12490--12505},
  year={2023},
  publisher={IEEE}
}

@article{fsdv2,
  title={Fsd v2: Improving fully sparse 3d object detection with virtual voxels},
  author={Fan, Lue and Wang, Feng and Wang, Naiyan and Zhang, Zhaoxiang},
  journal=PAMI,
  year={2024},
  publisher={IEEE}
}

@inproceedings{panoocc,
  title={Panoocc: Unified occupancy representation for camera-based 3d panoptic segmentation},
  author={Wang, Yuqi and Chen, Yuntao and Liao, Xingyu and Fan, Lue and Zhang, Zhaoxiang},
  booktitle={CVPR},
  year={2024}
}

@article{tan2025geocc,
  title={Geocc: Geometrically enhanced 3d occupancy network with implicit-explicit depth fusion and contextual self-supervision},
  author={Tan, Xin and Wu, Wenbin and Zhang, Zhiwei and Fan, Chaojie and Peng, Yong and Zhang, Zhizhong and Xie, Yuan and Ma, Lizhuang},
  journal={IEEE TITS},
  year={2025},
  publisher={IEEE}
}

@inproceedings{li2025occmamba,
  title={Occmamba: Semantic occupancy prediction with state space models},
  author={Li, Heng and Hou, Yuenan and Xing, Xiaohan and Ma, Yuexin and Sun, Xiao and Zhang, Yanyong},
  booktitle={CVPR},
  year={2025}
}

@inproceedings{huang2024selfocc,
  title={Selfocc: Self-supervised vision-based 3d occupancy prediction},
  author={Huang, Yuanhui and Zheng, Wenzhao and Zhang, Borui and Zhou, Jie and Lu, Jiwen},
  booktitle={CVPR},
  year={2024}
}

@article{lu2024lidar,
  title={LiDAR-camera continuous fusion in voxelized grid for semantic scene completion},
  author={Lu, Zonghao and Cao, Bing and Hu, Qinghua},
  journal=TCSVT,
  year={2024},
  publisher={IEEE}
}

@inproceedings{zhang2025visionpad,
  title={Visionpad: A vision-centric pre-training paradigm for autonomous driving},
  author={Zhang, Haiming and Zhou, Wending and Zhu, Yiyao and Yan, Xu and Gao, Jiantao and Bai, Dongfeng and Cai, Yingjie and Liu, Bingbing and Cui, Shuguang and Li, Zhen},
  booktitle={CVPR},
  year={2025}
}

@inproceedings{palladin2025self,
  title={Self-Supervised Sparse Sensor Fusion for Long Range Perception},
  author={Palladin, Edoardo and Brucker, Samuel and Ghilotti, Filippo and Narayanan, Praveen and Bijelic, Mario and Heide, Felix},
  booktitle={ICCV},
  year={2025}
}

@inproceedings{fan2025riocc,
  title={RIOcc: Efficient Cross-Modal Fusion Transformer with Collaborative Feature Refinement for 3D Semantic Occupancy Prediction},
  author={Fan, Baojie and Li, Xiaotian and Zhou, Yuhan and Jiang, Yuyu and Tian, Jiandong and Fan, Huijie},
  booktitle={CVPR},
  year={2025}
}

@inproceedings{zhu2025voxelsplat,
  title={Voxelsplat: Dynamic gaussian splatting as an effective loss for occupancy and flow prediction},
  author={Zhu, Ziyue and Wang, Shenlong and Xie, Jin and Liu, Jiang-jiang and Wang, Jingdong and Yang, Jian},
  booktitle=CVPR,
  year={2025}
}

@inproceedings{wang2025occrwkv,
  title={Occrwkv: Rethinking efficient 3d semantic occupancy prediction with linear complexity},
  author={Wang, Junming and Yin, Wei and Long, Xiaoxiao and Zhang, Xingyu and Xing, Zebin and Guo, Xiaoyang and Zhang, Qian},
  booktitle={ICRA},
  year={2025},
}

@inproceedings{guo2025sgformer,
  title={SGFormer: Satellite-Ground Fusion for 3D Semantic Scene Completion},
  author={Guo, Xiyue and Hu, Jiarui and Hu, Junjie and Bao, Hujun and Zhang, Guofeng},
  booktitle={CVPR},
  year={2025}
}

@article{li2025omninwm,
  title={OmniNWM: Omniscient Driving Navigation World Models},
  author={Li, Bohan and Ma, Zhuang and Du, Dalong and Peng, Baorui and Liang, Zhujin and Liu, Zhenqiang and Ma, Chao and Jin, Yueming and Zhao, Hao and Zeng, Wenjun and others},
  journal={arXiv:2510.18313},
  year={2025}
}

@article{zuo2025dvgt,
  title={DVGT: Driving Visual Geometry Transformer},
  author={Zuo, Sicheng and Xie, Zixun and Zheng, Wenzhao and Xu, Shaoqing and Li, Fang and Jiang, Shengyin and Chen, Long and Yang, Zhi-Xin and Lu, Jiwen},
  journal={arXiv:2512.16919},
  year={2025}
}

@article{kerbl20233d,
  title={3D Gaussian splatting for real-time radiance field rendering.},
  author={Kerbl, Bernhard and Kopanas, Georgios and Leimk{\"u}hler, Thomas and Drettakis, George},
  journal=TOG,
  volume={42},
  number={4},
  pages={139--1},
  year={2023}
}

@inproceedings{ye2025gs,
  title={Gs-occ3d: Scaling vision-only occupancy reconstruction with gaussian splatting},
  author={Ye, Baijun and Qin, Minghui and Zhang, Saining and Gong, Moonjun and Zhu, Shaoting and Zhao, Hao and Zhao, Hang},
  booktitle=ICCV,
  year={2025}
}

@inproceedings{yu2025language,
  title={Language driven occupancy prediction},
  author={Yu, Zhu and Pang, Bowen and Liu, Lizhe and Zhang, Runmin and Li, Qiang and Cao, Si-Yuan and Luo, Maochun and Chen, Mingxia and Yang, Sheng and Shen, Hui-Liang},
  booktitle=ICCV,
  year={2025}
}

@inproceedings{zheng2024occworld,
  title={Occworld: Learning a 3d occupancy world model for autonomous driving},
  author={Zheng, Wenzhao and Chen, Weiliang and Huang, Yuanhui and Zhang, Borui and Duan, Yueqi and Lu, Jiwen},
  booktitle=ECCV,
  year={2024},
}

@inproceedings{gan2025gaussianocc,
  title={Gaussianocc: Fully self-supervised and efficient 3d occupancy estimation with gaussian splatting},
  author={Gan, Wanshui and Liu, Fang and Xu, Hongbin and Mo, Ningkai and Yokoya, Naoto},
  booktitle=ICCV,
  year={2025}
}

@inproceedings{chambon2025gaussrender,
  title={Gaussrender: Learning 3d occupancy with gaussian rendering},
  author={Chambon, Loick and Zablocki, Eloi and Boulch, Alexandre and Chen, Mickael and Cord, Matthieu},
  booktitle=ICCV,
  year={2025}
}

@inproceedings{zuo2025gaussianworld,
  title={Gaussianworld: Gaussian world model for streaming 3d occupancy prediction},
  author={Zuo, Sicheng and Zheng, Wenzhao and Huang, Yuanhui and Zhou, Jie and Lu, Jiwen},
  booktitle=CVPR,
  year={2025}
}

@inproceedings{jiang2025gausstr,
  title={Gausstr: Foundation model-aligned gaussian transformer for self-supervised 3d spatial understanding},
  author={Jiang, Haoyi and Liu, Liu and Cheng, Tianheng and Wang, Xinjie and Lin, Tianwei and Su, Zhizhong and Liu, Wenyu and Wang, Xinggang},
  booktitle=CVPR,
  year={2025}
}

@inproceedings{zheng2025bi,
  title={Bi-Stream Knowledge Transfer for Semi-Supervised 3D Point Cloud Object Detection},
  author={Zheng, Jilai and Tang, Pin and Ren, Xiangxuan and Wang, Zhongdao and Ma, Chao},
  booktitle=ICRA,
  year={2025},
}

@article{ren2025litefusion,
  title={LiteFusion: Taming 3D Object Detectors from Vision-Based to Multi-Modal with Minimal Adaptation},
  author={Ren, Xiangxuan and Wang, Zhongdao and Tang, Pin and Wang, Guoqing and Zheng, Jilai and Ma, Chao},
  journal={arXiv:2512.20217},
  year={2025}
}

@inproceedings{zhu2021cylindrical,
  title={Cylindrical and asymmetrical 3d convolution networks for lidar segmentation},
  author={Zhu, Xinge and Zhou, Hui and Wang, Tai and Hong, Fangzhou and Ma, Yuexin and Li, Wei and Li, Hongsheng and Lin, Dahua},
  booktitle={CVPR},
  year={2021}
}

@inproceedings{bai2022transfusion,
  title={Transfusion: Robust lidar-camera fusion for 3d object detection with transformers},
  author={Bai, Xuyang and Hu, Zeyu and Zhu, Xinge and Huang, Qingqiu and Chen, Yilun and Fu, Hongbo and Tai, Chiew-Lan},
  booktitle={CVPR},
  year={2022}
}

@inproceedings{liu2023uniseg,
  title={Uniseg: A unified multi-modal lidar segmentation network and the openpcseg codebase},
  author={Liu, Youquan and Chen, Runnan and Li, Xin and Kong, Lingdong and Yang, Yuchen and Xia, Zhaoyang and Bai, Yeqi and Zhu, Xinge and Ma, Yuexin and Li, Yikang and others},
  booktitle={ICCV},
  year={2023}
}

@inproceedings{li2023mseg3d,
  title={Mseg3d: Multi-modal 3d semantic segmentation for autonomous driving},
  author={Li, Jiale and Dai, Hang and Han, Hao and Ding, Yong},
  booktitle={CVPR},
  year={2023}
}

@inproceedings{zheng2024genad,
  title={Genad: Generative end-to-end autonomous driving},
  author={Zheng, Wenzhao and Song, Ruiqi and Guo, Xianda and Zhang, Chenming and Chen, Long},
  booktitle=ECCV,
  year={2024},
}
}

\vspace{11pt}

\vfill

\end{document}